%% file: main.tex
\documentclass[10pt,twocolumn,letterpaper]{article}
\usepackage{multirow}
\usepackage[table]{xcolor} %
\usepackage{bm} 
\usepackage{cvpr}              %
\usepackage[ruled,vlined]{algorithm2e}
\usepackage{placeins}
\usepackage[table]{xcolor}
\usepackage{booktabs}
\usepackage{subcaption}

\usepackage{adjustbox}
\usepackage{etoc}
\usepackage{makecell}
\usepackage[accsupp]{axessibility}  %

\usepackage{tocloft}

\input{preamble}
\definecolor{cvprblue}{rgb}{0.21,0.49,0.74}
\usepackage[pagebackref,breaklinks,colorlinks,allcolors=cvprblue]{hyperref}

\def\confName{CVPR}
\def\confYear{2026}

\title{Space-Time Forecasting of Dynamic Scenes\\ with Motion-aware Gaussian Grouping}

\author{
Junmyeong Lee\textsuperscript{1*} \quad
Hoseung Choi\textsuperscript{1*} \quad
Minsu Cho\textsuperscript{1,2}
\\
{\tt\small \{junmyeong.lee, hs.choi, mscho\}@postech.ac.kr}
\\[0.5em]
\textsuperscript{1} Pohang University of Science and Technology (POSTECH) \quad
\textsuperscript{2} RLWRLD
}

\begin{document}
\let\origaddcontentsline\addcontentsline
\renewcommand{\addcontentsline}[3]{}
\twocolumn[{%
\renewcommand\twocolumn[1][]{#1}%
\maketitle
\begin{center}
\centering
\captionsetup{type=figure}
\vspace{-5mm}
\includegraphics[width=\textwidth]{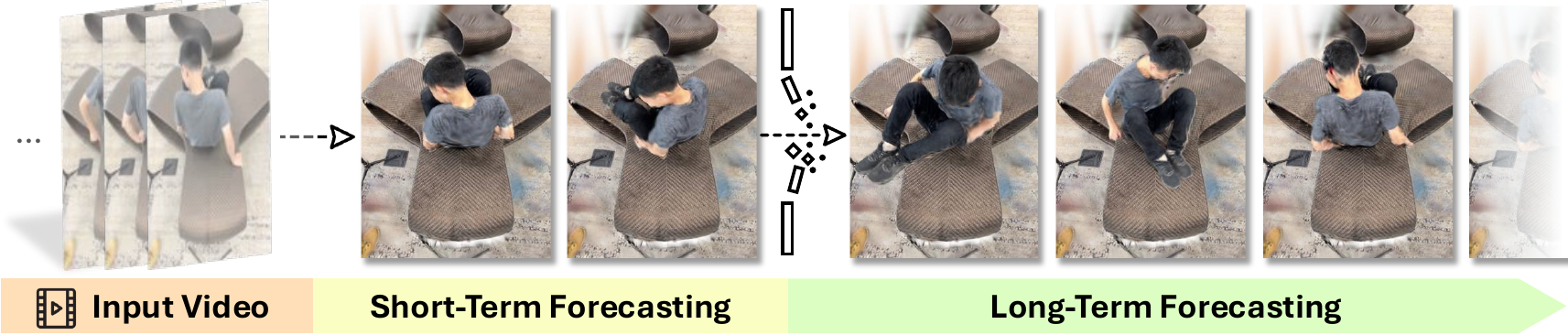}
\captionof{figure}{
\textbf{Motion Group-aware Gaussian Forecasting (\textbf{MoGaF})}. Our method predicts future frames of a dynamic input video by constructing object-level groups of Gaussian splatting with distinct motion patterns and leveraging these motion group structures in both space and time. The proposed method enables long-term, high-fidelity predictions on real-world videos with complex dynamics.
}

\end{center}%
}]
\def\thefootnote{*}\footnotetext{Equal contribution.}

\input{sec/0_abstract}    
\input{sec/1_intro}

\input{sec/2_related}

\input{sec/3_method}

\input{sec/4_experiments}
\input{sec/5_conclusion}
\clearpage
\input{sec/6_acknowledgement}
{
    \small
    \bibliographystyle{ieeenat_fullname}
    \bibliography{main}
}

\clearpage
\setcounter{page}{1}
\setcounter{section}{0}
\maketitlesupplementary

\renewcommand{\thesection}{\Alph{section}}
\renewcommand{\thesubsection}{\Alph{section}.\arabic{subsection}}

\renewcommand{\theHsection}{supple.\Alph{section}}
\renewcommand{\theHsubsection}{supple.\Alph{section}.\arabic{subsection}}

\let\addcontentsline\origaddcontentsline

{
  \hypersetup{linkcolor = black}
  \renewcommand{\cftsecleader}{}
  \renewcommand{\cftsecafterpnum}{}
  \cftpagenumbersoff{section}
  \cftpagenumbersoff{subsection}
  \tableofcontents
  \thispagestyle{empty}
}

\input{supple_sec/2_grouping}
\input{supple_sec/1_implementation}

\input{supple_sec/3_additional_results}

\input{supple_sec/4_limitations}
\input{supple_sec/5_figures}

\end{document}

%% file: sec/0_abstract.tex
\begin{abstract}
Forecasting dynamic scenes remains a fundamental challenge in computer vision, as limited observations make it difficult to capture coherent object-level motion and long-term temporal evolution.
We present \textbf{Motion Group-aware Gaussian Forecasting (\oursname)}, a framework for long-term scene extrapolation built upon the 4D Gaussian Splatting representation.
{\oursname} introduces motion-aware Gaussian grouping and group-wise optimization to enforce physically consistent motion across both rigid and non-rigid regions, yielding spatially coherent dynamic representations.
Leveraging this structured space-time representation, a lightweight forecasting module predicts future motion, enabling realistic and temporally stable scene evolution.
Experiments on synthetic and real-world datasets demonstrate that {\oursname} consistently outperforms existing baselines in rendering quality, motion plausibility, and long-term forecasting stability. Our project page is available at \url{https://slime0519.github.io/mogaf}
\end{abstract}

%% file: sec/1_intro.tex
\vspace{-4mm}
\section{Introduction}
Reconstructing dynamic environments from limited observations is a fundamental challenge in computer vision.
Recent advances such as 3D Gaussian Splatting~\cite{kerbl20233d} have enabled dynamic scene reconstruction from everyday videos and multi-view captures.
However, most existing methods still focus on \textit{interpolation} of observed dynamics, 
while \textit{extrapolation} remains largely unexplored.
The ability to extrapolate scene dynamics is increasingly important for emerging applications such as robotic behavior planning and autonomous driving, where anticipating unobserved future motion is critical for decision making.
To bridge this gap, we aim to forecast future scene dynamics by leveraging optimized scene representation from available observations.

Earlier approaches to dynamic scene extrapolation~\cite{wu2020future, wu2022optimizing, girdhar2021anticipative, kwon2019predicting, lu2017flexible} formulate it as video prediction, generating future frames from past observations.
Robotic planning models~\cite{wu2024ivideogpt, assran2025v, hafner2019dream} adopt a similar idea, using generative models to predict future visual states from current inputs.
However, these 2D-based methods are restricted to fixed-view generation and often produce geometrically inconsistent results in complex scenes~\cite{ming2024survey}.
In contrast, 3D reconstruction–based methods~\cite{mildenhall2021nerf, kerbl20233d} model time-dependent motion within explicit 3D scene representation.
However, these approaches remain fundamentally \textit{interpolative}, offering little \textit{extrapolative} capability beyond the observed timestamps.
As a result, their predicted dynamics often degrade when extended outside the training range.
\cite{zhao2024gaussianprediction} incorporates explicit motion modeling to improve short-term prediction but still struggles with long-term forecasting.

Forecasting long-term dynamics remains challenging due to limitations in both representation and architecture.
At the representation level, prior methods~\cite{wang2025shape, lei2025mosca} shows that Gaussian primitives often move independently, causing spatially incoherent motion that accumulates over time.
At the architectural level, existing extrapolation model~\cite{zhao2024gaussianprediction} are typically short-term predictors that produce frozen or collapsing trajectories during long rollouts.

To address these issues, we propose \textbf{Motion Group-aware Gaussian Forecasting (\oursname)},
a unified framework for physically consistent long-term extrapolation based on 4D Gaussian Splatting (4DGS).
{\oursname} introduces (1) motion-aware Gaussian grouping that identifies coherent motion groups, (2) group-wise optimization that enforces consistent motion for rigid and non-rigid regions,
and (3) lightweight motion forecasting to extrapolate future dynamics from the 4DGS representation.
Integrating these components into a single pipeline yields spatially coherent and temporally stable scene evolution,
enabling realistic long-term forecasting that surpasses baselines in both visual quality and motion plausibility.

\noindent Our contributions are summarized as follows:
\begin{itemize}
\item We present \textbf{\oursname}, a framework that integrates object-level motion modeling into dynamic Gaussian Splatting for long-term scene extrapolation.
\item Motion-aware Gaussian grouping, group-wise optimization, and a lightweight forecaster that jointly yield physically consistent and temporally coherent scene evolution.
\item Extensive experiments on synthetic and real-world datasets demonstrate that \oursname{} surpasses existing baselines in the visual quality of future frame synthesis.
\end{itemize}

%% file: sec/2_related.tex
\section{Related work}
\label{sec:related_work}

\subsection{Dynamic Scene Reconstruction}
Dynamic scene reconstruction aims to recover time-varying 3D structures and appearance from observations. 
Early approaches represent deformable geometry with explicit primitives such as meshes, point clouds, or parametric models.
Inspired by the success of NeRF~\cite{mildenhall2021nerf}, neural field--based methods~\cite{pumarola2021d, park2021nerfies, park2021hypernerf} extend implicit representation to model temporal deformation fields.
Another line of work~\cite{li2022neural, fridovich2023k, cao2023hexplane} encode dynamics implicitly within feature spaces.

3D Gaussian Splatting (3DGS)~\cite{kerbl20233d} offers an explicit repsentation, 
enabling real-time rendering through volumetric Gaussian primitives and an efficient rasterization pipeline. Building on 3DGS, several methods learn per-Gaussian trajectories or factorized motion bases to obtain 4D representation~\cite{luiten2024dynamic, wu20244d, kratimenos2024dynmf, yang2023real, yang2024deformable}. 
However, many of these approaches assume limited object and camera motion, often leading to reconstruction failures and artifacts in highly dynamic scenes.
Recent efforts~\cite{wang2025shape, lei2025mosca, luo2025instant4d4dgaussiansplatting} mitigate these issues by introducing improved initialization and optimization strategies that leverage data-driven priors such as dynamic masks~\cite{yang2023track}, dense point tracking~\cite{karaev2024cotracker, doersch2023tapir}, and depth estimation~\cite{yang2024depth}.
Building on this direction, {\oursname} groups Gaussians into coherent motion units and enforces group-wise motion consistency during optimization.

\subsection{3D Gaussian Segmentation}

With the emergence of reliable 2D foundation models for segmentation~\cite{ravi2024sam, cen2025segment, kirillov2023segment}, 
recent work has explored extending their prior knowledge to obtain 3D-consistent segmentation. 
Some methods~\cite{chen2023interactive, kim2024garfield, zhou2024feature} distill 2D foundation-model features into neural-field representation. 
Another stream of research lifts 2D mask predictions into 3D for multi-view segmentation without retraining. 
However, models such as SAM~\cite{ravi2024sam} do not preserve instance identities across views.
This limitation has led to approaches~\cite{cen2025segment, ye2024gaussian, dou2024learning} that associate identities via cross-view cues such as user interaction or video object tracking~\cite{cheng2023tracking}. 
Gaga~\cite{lyu2024gaga} further proposes a 3D-aware mask association pipeline using a Gaussian memory bank and IoU-based merging. 
Although effective for static scenes, these methods remain difficult to extend to dynamic settings due to temporal object motion.

In this work, we propose a segmentation method for 4DGS representation by extending the 3DGS-based segmentation pipeline~\cite{lyu2024gaga} to dynamic scenes. 
This extension yields spatiotemporally coherent Gaussian groups.
Our grouping strategy relies on alternating spatiotemporal region growing and keyframe-based Gaussian registration. 
This procedure is simple yet effective, enabling reliable grouping on an optimized 4DGS representation.

\subsection{Dynamic Scene Forecasting}
To extrapolate scene dynamics, several prior studies~\cite{wang2018eidetic, liu2018dyan} focused on pixel-space video prediction using convolutional or recurrent architectures, where future frames are generated directly in the image domain. 
Following this direction, generative models~\cite{yan2021videogpt, ye2022vptr} leverage transformer-based architectures for scalable video forecasting, and subsequent work~\cite{gupta2022maskvit, villar2023object} further incorporates masked video modeling and self-supervised pretraining.

Building on recent progress in 3D reconstruction, another line of work predicts future dynamics directly in 3D space.  
This direction has recently been explored in the 4DGS setting as well, %
including GaussianPrediction (GSPred)~\cite{zhao2024gaussianprediction}, which forecasts keypoint motion using a graph-based network and propagates it to all Gaussians, and a concurrent work~\cite{wang2025ode} that models continuous Gaussian motion via neural differential equations.  
Despite these advances, existing 3D-based models remain effective only for short-term prediction and often degrade significantly when extended to longer horizons.  
To address this limitation, our method performs object-wise forecasting and preserves both rigid and non-rigid geometry, enabling structurally consistent and high-fidelity rendering.

%% file: sec/3_method.tex
\section{Preliminary}
\label{subsec: method_gaussian_splatting}
\paragraph{Gaussian Splatting.}
3DGS~\cite{kerbl20233d} represents scene geometry using a set of Gaussian primitives and achieves real-time, high-fidelity rendering through efficient tile-based rasterization.
Following \cite{wang2025shape}, each Gaussian $g \in \mathcal{G}$ is parameterized in a canonical space by
$\{ \bm{\mu}, \bm{R}, \bm{s}, o, \bm{c} \}$,
which specify its mean, rotation, scale, opacity, and color.
Here, $\bm{\mu} \in \mathbb{R}^3$ and $\bm{R} \in SO(3)$ denote the 3D mean and rotation, 
while $\bm{s} \in \mathbb{R}^3$, $o \in \mathbb{R}$, and $\bm{c} \in \mathbb{R}^3$ represent its scale, opacity, and color.

According to \cite{ewasplatting}, Gaussians are projected onto the image plane by approximating their 2D means and covariances as follows:
\begin{equation}
{\bm{\mu}}^{2D} = \bm{\Pi}(\bm{K}\bm{E}{\bm{\mu}}), \
\quad
{\bm{\varSigma}}^{2D} = \bm{J}\bm{E}{\bm{\varSigma}}\bm{E}^\top\bm{J}^\top,
\end{equation}
where ${\bm{\varSigma}}$ and ${\bm{\varSigma}}^{2D}$ denote the 3D and 2D covariance matrices, respectively.
$\bm{J}$ is the Jacobian of the affine approximation of the projective transformation, and $\bm{K}$ and $\bm{E}$ are the intrinsic and extrinsic matrices of the camera. $\bm{\Pi}$ denotes the perspective projection from 3D points to the image plane.
Each covariance matrix is decomposed into a rotation matrix $\bm{R}$ and a scaling matrix $\bm{S}$ such that
$$\bm{\varSigma} = \bm{R}\bm{S}\bm{S}^\top\bm{R}^\top.$$
In addition, each Gaussian includes an opacity $\alpha \in \mathbb{R}$ and spherical harmonics (SH) coefficients $\bm{c} \in \mathbb{R}^{(L+1)^2}$ to represent view-dependent color.
Thus, the final color of pixel \( \bm{x}_p \) is computed as:
\begin{equation}
    C_p = \sum_{i=1}^{N} c_i \alpha_i T_i \mathcal{N}\left(\bm{x}_p|\bm{\mu}^{2D},\bm{\varSigma}^{2D} \right),
\end{equation}
where \(T_i = \prod_{j=1}^{i-1} \left(1-\alpha_j \mathcal{N}\left(\bm{x}_p|\bm{\mu}^{2D},\bm{\varSigma}^{2D}\right) \right)\), and \( c_i \) and \( \alpha_i \) represent color and opacity associated with each 3D Gaussian, respectively.
\vspace{-3mm}
\paragraph{Extension for Dynamic Scene.}Then, the time-dependent motion parameters ${\deformed{\bm{\mu}}{t}}$ and ${\deformed{\bm{R}}{t}}$ 
are obtained by composing the canonical Gaussian parameters—mean ${\canonical{\bm{\mu}}}$, 
rotation ${\canonical{\bm{R}}}$—with the motion transformation 
${\motionbasis{T}{c}{t} = [\motionbasis{R}{c}{t} \mid \motionbasis{t}{c}{t}] \in SE(3)}$, defined as:
\begin{equation}
    \deformed{\bm{\mu}}{t} = \motionbasis{R}{c}{t}\canonical{\bm{\mu}} + \motionbasis{t}{c}{t}, \quad
    \deformed{\bm{R}}{t} = \motionbasis{R}{c}{t}\canonical{\bm{R}}.
\end{equation}
For each timestep ${t}$, we represent Gaussian motion using a set of ${B}$ motion bases 
${\{\motionbasissuper{T}{b}{c}{t}\}_{b=1}^B}$ shared across all timesteps 
${t = 0, 1, \dots, T}$. 
Each Gaussian is assigned a blend weight (i.e., motion coefficient) 
${\bm{w} \in \mathbb{R}^B}$ to combine these bases and represent its motion. 
At timestep ${t}$, the composed motion transformation is computed as:
\begin{equation}
    \motionbasis{T}{c}{t} = \sum_{b=1}^B w^{(b)} \motionbasissuper{T}{b}{c}{t},    
\end{equation}
where ${\lVert \bm{w} \rVert = 1}$. 
During optimization, both the motion bases ${\{\motionbasissuper{T}{b}{c}{t}\}_{b=1}^B}$ 
and the motion coefficient ${\bm{w}^b}$ are jointly optimized 
with the other Gaussian properties.
The remaining rendering process follows the 3DGS formulation, 
using the deformed rotation ${\deformed{\bm{R}}{t}}$ 
and mean vector ${\deformed{\bm{\mu}}{t}}$.

\section{Methods}
\begin{figure*}[ht!]
\centering
\includegraphics[width=\textwidth]{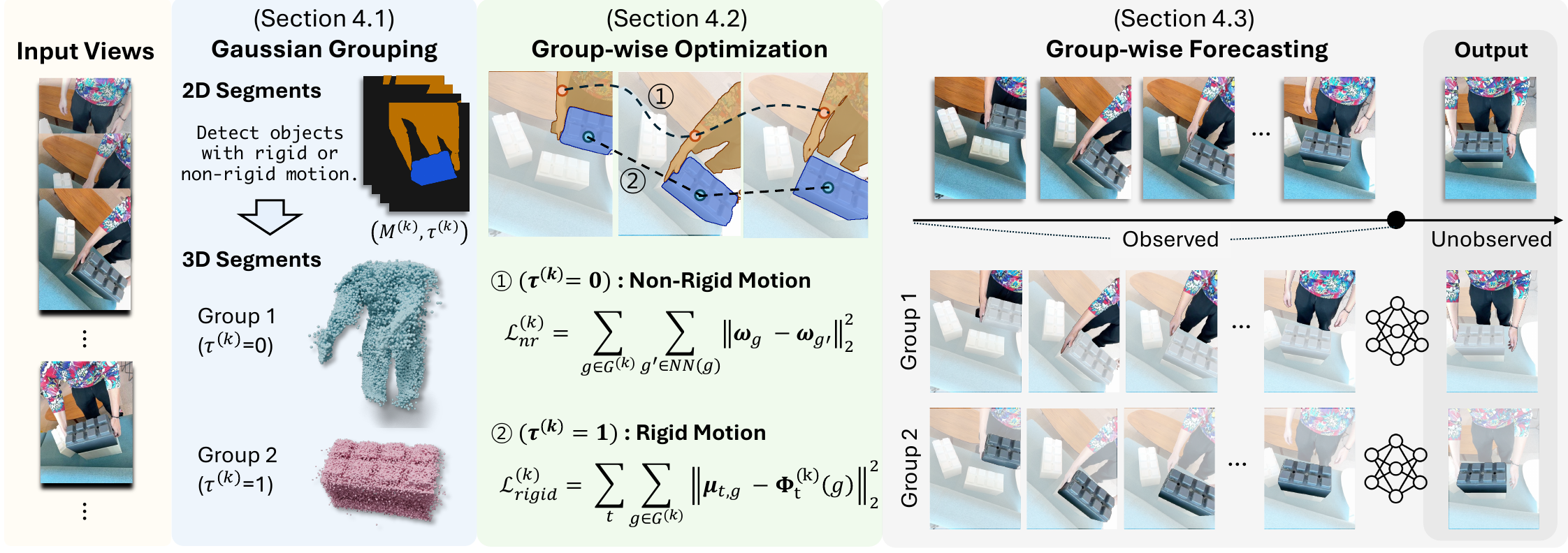}
\vspace{-6mm}
\caption{\textbf{Overall pipeline of \oursname.}
Given a video, \oursname{} generates future frames of the scene. 
To achieve realistic forecasting, our method builds on 4DGS representation and proceeds as follows:
(1) \textbf{Gaussian Grouping:} Gaussians are clustered into motion-consistent object groups, with each group labeled as rigid or non-rigid using grounded 2D segmentation.
(2) \textbf{Group-wise Optimization:} Grouped Gaussians are refined with rigidity-aware motion constraints: rigid groups are guided by a shared $SE(3)$ transform, while non-rigid groups are regularized with local motion smoothness. 
(3) \textbf{Group-wise Forecasting:} For each group, a lightweight Transformer-based forecaster extrapolates Gaussian trajectories beyond the observed frames, enabling rendering at novel viewpoints for future timesteps.}
\label{fig:main-figure}
\vspace{-5mm}
\end{figure*}

Our method, \oursname, takes as input a casually captured dynamic video 
${\{I_t \in \mathbb{R}^{H \times W \times 3}\}_{t=1}^T}$ and forecasts scene dynamics 
by rendering novel frames for unseen timesteps $({t > T})$. 
Unlike previous methods~\cite{zhao2024gaussianprediction, wang2025ode}, 
our goal is to achieve realistic and physically consistent long-term forecasting at the scene level.  

We address the limitations of dynamic 3D representation by reconstructing scene dynamics through a constrained optimization process that enforces physically consistent motion across time.
{\oursname} models object-level motion explicitly within the dynamic Gaussian splatting framework.

The overall pipeline consists of three stages:
(i) motion-aware Gaussian grouping, which identifies coherent motion groups and labels each group as either rigid or non-rigid (Section~\ref{sec:gs-grouping});
(ii) group-wise constrained optimization, which applies type-specific motion regularization to produce a physically structured 4D representation with object-level consistency (Section~\ref{sec:group-optimization}); and
(iii) group-wise long-term motion forecasting, where a temporal prediction module extrapolates future motion trajectories and renders temporally coherent scene evolution (Section~\ref{sec:motion-forecasting}).
An overview of our framework is illustrated in Figure~\ref{fig:main-figure}.

\subsection{Motion-aware Gaussian Grouping}
\label{sec:gs-grouping}

\begin{figure}[t]
\centering
\includegraphics[width=0.9\columnwidth]{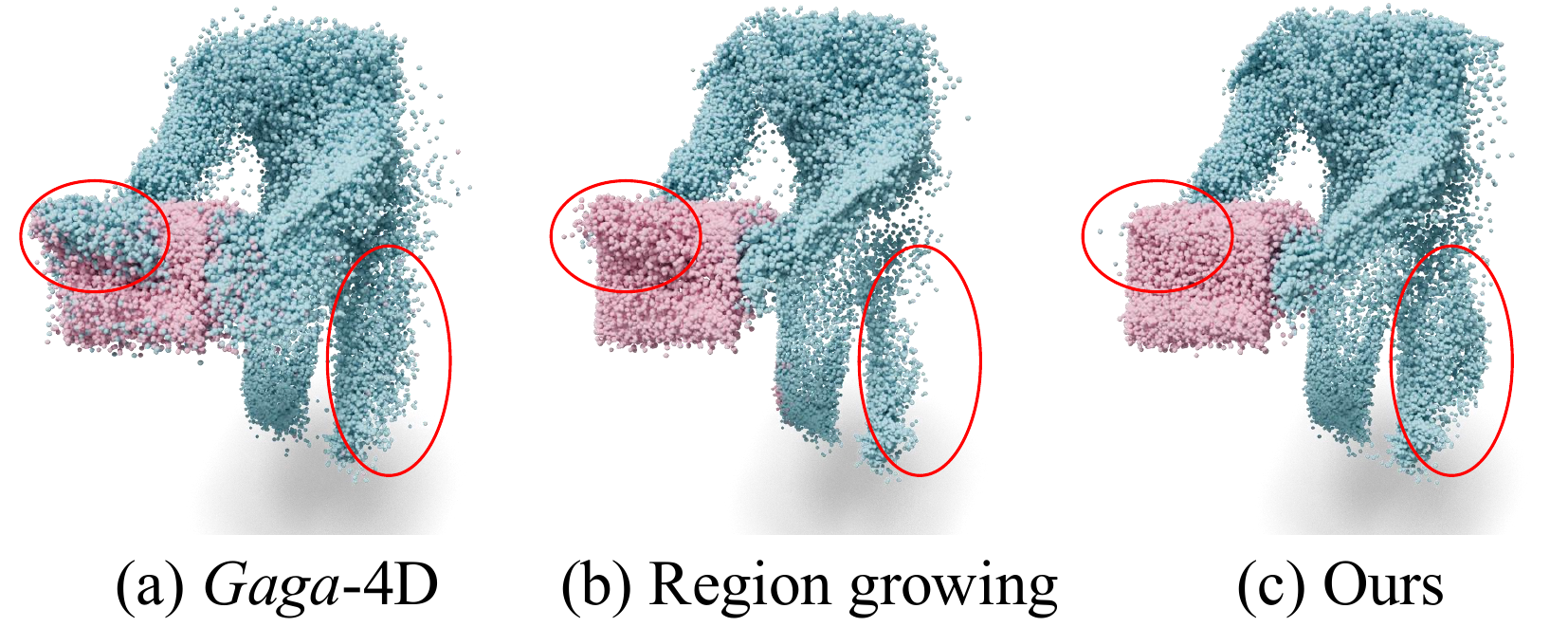}
\vspace{-2mm}
\caption{\textbf{Result of Gaussian grouping.} 
Compared to a (a) simple extension of 3DGS grouping~\cite{lyu2024gaga} and (b) single-frame mask–based region growing, 
our hybrid approach produces complete and reliable motion-aware Gaussian groups.}
\label{fig:grouping_comparison}
\vspace{-4mm}
\end{figure}

Given the frames $\{I_t\}_{t=1}^T$ and the reconstructed 4DGS representation, 
we group Gaussians into distinct groups with coherent motion and label each group as \textit{rigid} or \textit{non-rigid} 
to support subsequent motion-constrained optimization and forecasting. 
Inspired by \cite{lyu2024gaga},
we design a memory-bank mechanism adapted to the optimized 4DGS representation. 
Unlike prior 3DGS-based grouping methods, our module handles dynamic representation 
and explicitly distinguishes motion types of groups.
Given $K$ object groups, we represent them as 
$\motiongroup{\mathcal{M}}{k} = (\motiongroup{G}{k},\,\motiongroup{\tau}{k})$, 
where $\motiongroup{G}{k}$ is the set of Gaussians in the $k$-th group 
and $\motiongroup{\tau}{k}\!\in\!\{0,1\}$ indicates its motion type 
($0$ for \textit{non-rigid} and $1$ for \textit{rigid}). 
An overview of the grouping algorithm is shown in Algorithm~\ref{alg:gaussian_grouping_compact}.  
Detailed analyses of our Gaussian grouping are provided in the Supplementary Material (\Cref{sup_sec:grouping_details,sup_sec:grouping-analysis}).
\vspace{-4mm}
\paragraph{Leverage 2D Segmentation Prior.}
Similar to previous methods~\cite{lyu2024gaga, ye2024gaussian}, we first obtain 2D masks for the input video.
In particular, we use a grounded segmentation model~\cite{ren2024grounding} to simultaneously produce ${K}$ object masks and their corresponding rigidity labels ${\motiongroup{\tau}{k}}$.
Following \cite{lyu2024gaga}, we then identify Gaussians that contribute to rendering each masked region 
and extract these front Gaussians along the viewing direction. 

\vspace{-4mm}
\paragraph{Simple Extension of Static Gaussian Grouping.}
A straightforward way to extend static Gaussian grouping~\cite{lyu2024gaga} to dynamic scenes 
is to associate each 2D mask $\motiongroup{M}{k}_t$ with the set of deformed Gaussians 
whose projected positions fall inside the mask:
\begin{equation}
    \motiongroup{G}{k}_t = \{\, g \in \mathcal{G} \mid \text{Proj}({g}_{t}) \in \motiongroup{M}{k}_t \,\},
\end{equation}
where ${g}_{t}$ denotes the deformed Gaussian state at timestep $t$. We then update each Gaussian group $\motiongroup{G}{k} \in \motiongroup{\mathcal{M}}{k}$ in the memory bank 
by merging $\motiongroup{G}{k}_t$ when sufficient spatial overlap is observed.  
However, because this association relies solely on deformed spatial positions of Gaussians, it often leads to misgrouping (Figure~\ref{fig:grouping_comparison}a), especially under occlusion or when Gaussians from different objects overlap.

\vspace{-3mm}
\paragraph{Gaussian Grouping via Iterative Region Growing.}
To overcome the limitations of this simple algorithm,
we adopt an iterative region-growing strategy that alternates between keyframe-based seeding and feature-space expansion.
At each keyframe, we project the deformed Gaussians and select the front-most ones that contribute to rendering each mask as reliable seeds $\motiongroup{G}{k}_t$.
These seeds are merged into their corresponding motion groups $\motiongroup{G}{k}$ in the memory bank, which then serve as the starting points for region growing.
During region growing, each Gaussian $g \in \mathcal{G}$ is represented by a compact spatiotemporal feature
$\bm{f}_g = [\canonicalsub{\mu}{c,g},\bm{w}’_g,]$,
where $\canonicalsub{\mu}{c,g}$ denotes its canonical-space mean and $\bm{w}’_g$ encodes its PCA-reduced motion coefficient.
Using these features, each group $\motiongroup{G}{k}$ is expanded by aggregating nearby Gaussians that satisfy
$| \bm{f}_g - \bm{f}_{g’} | < \epsilon_r$.
The updated groups are stored in the memory bank, and the process continues at the next keyframe.

Through this alternating loop of front-Gaussian seeding and feature-space region growing, 
our method effectively captures the spatiotemporal characteristics of the dynamic Gaussian representation,
resulting in robust and accurate motion grouping.
As shown in Figure~\ref{fig:grouping_comparison}, our approach produces reliable object-level motion groups with highly accurate assignments, outperforming competing methods.

\begin{algorithm}[t]
\caption{\textit{Motion-aware Gaussian Grouping}}
\label{alg:gaussian_grouping_compact}
\KwIn{Video frames $\{I_t\}_{t=1}^T$, segmentation model $\mathcal{S}$, dynamic Gaussian set $\mathcal{G}$ with deformed Gaussian states $\{ g_t \}_{t=1}^T$}
\KwOut{Motion groups $\{\motiongroup{\mathcal{M}}{k}=(\motiongroup{G}{k},\,\motiongroup{\tau}{k})\}_{k=1}^K$}

\BlankLine
\textbf{Stage 1: Init at the First Keyframe}

Obtain $\{\motiongroup{M}{k}_t,\,\motiongroup{\tau}{k}_{\text{mask}}\} \leftarrow \mathcal{S}(\{I_t\})$ \\
Select first keyframe $t_k$, \\
\ForEach{$k=1,\dots,K$}{ 
  $\motiongroup{G}{k} \leftarrow \{\, g\in\mathcal{G} \mid \text{Proj}(g_{t_k}) \in \motiongroup{M}{k}_{t_k}\}$ \\
  $\motiongroup{\tau}{k} \leftarrow \motiongroup{\tau}{k}_{\text{mask}}$
}
Define features for all $g$: $\bm{f}_g = [\,\canonicalsub{\mu}{c,g},\,\bm{w}'_g\,]$

\BlankLine
\textbf{Stage 2: Iterative Region Growing} \\
\ForEach{keyframe $t$ in stride order}{

  \small{\tcp{(A) Feature-space region growing}}
  \Repeat{\textit{no group changes}}{
    $U \leftarrow \mathcal{G}\setminus \big(\bigcup_{k} \motiongroup{G}{k}\big)$ \\
    \ForEach{$k=1,\dots,K$}{
      $\epsilon_k \leftarrow \alpha \cdot \mathop{\mathrm{mean}}\limits_{g\in \motiongroup{G}{k}} \mathrm{KNN}_K(\bm{f}_g)$ \\
      $N_k \leftarrow \{\, g' \in U \mid \mathop{\min}\limits_{g'\in \motiongroup{G}{k}} \|\bm{f}_g-\bm{f}_{g'}\| < \epsilon_k \,\}$ \\
      $\motiongroup{G}{k} \leftarrow \motiongroup{G}{k} \cup N_k$ \\
      $U \leftarrow U \setminus N_k$
    }
  }
  \small{\tcp{(B) Seeding \& merge at keyframe $t$}}
  \ForEach{$k=1,\dots,K$}{
    $\motiongroup{G}{k}_t \leftarrow \{\, g \mid \text{Proj}(g_{t}) \in \motiongroup{M}{k}_t\}$ \\
    $\motiongroup{G}{k} \leftarrow \motiongroup{G}{k} \cup \motiongroup{G}{k}_t$
  }
}
\end{algorithm}

\subsection{Group-wise Gaussian Motion Optimization}
\label{sec:group-optimization}
After grouping, we further optimize the 4DGS representation by incorporating the object group information. 
We introduce group-wise motion constraints guided by the rigidity flag ${\motiongroup{\tau}{k}}$, 
which adaptively applies distinct motion regularization strategies to rigid and non-rigid object groups. 
\vspace{-8mm}
\paragraph{Rigid Motion Anchoring Loss.}
We perform constrained optimization to refine the grouped 4DGS representation by adaptively applying motion constraints 
to each Gaussian ${g \in \motiongroup{G}{k}}$ according to the group motion type. For a group labeled as rigid (${ \motiongroup{\tau}{k}= 1 }$), 
all Gaussians within the group are constrained to share a single $SE(3)$ transformation 
${\bm{\Phi}_{t}^{(k)} = [\, \rigidmotion{R}{k}{c}{t} \mid \rigidmotion{t}{k}{c}{t}\,]}$.
We define the group-level rigid transformation as a function acting on a Gaussian $g \in G_k$ 
that maps its canonical-space mean to the deformed space at time $t$:
\begin{equation}
    \bm{\Phi}_{t}^{(k)}(g)
    =
    \rigidmotion{R}{k}{c}{t}\canonicalsub{\mu}{c,g}
    + \rigidmotion{t}{k}{c}{t}.
\end{equation}
This operator ensures that all Gaussians within the same rigid group undergo a consistent $SE(3)$ motion, 
preserving intra-group spatial structure during optimization.
The rigid-body loss then measures the deviation between the learned per-Gaussian motion and the group-level rigid transform:
\begin{equation}
    \mathcal{L}_{\text{rigid}}^{(k)} =
    \sum_{t}\sum_{g \in \motiongroup{G}{k}}
    \big\|
        \deformed{\mu}{t,g}
        - \bm{\Phi}_{t}^{(k)}(g)
    \big\|_2^2.
\end{equation}

\paragraph{Non-Rigid Motion Smoothness Loss.}
For non-rigid groups (${ \motiongroup{\tau}{k} = 0 }$), 
each Gaussian ${g \in \mathcal{G}}$ is assigned a learnable motion coefficient 
${\bm{w}_g \in \mathbb{R}^{B}}$ that governs its deformation behavior over time. 
Following \cite{wang2025shape}, we promote locally coherent motion in the canonical space 
by applying a spatial smoothness regularization defined as
\begin{equation}
    \mathcal{L}_{\text{nr}}^{(k)}
    =
    \sum_{g \in \motiongroup{G}{k}}
    \sum_{g' \in \text{NN}(g)}
    \|\bm{w}_g - \bm{w}_{g'}\|_2^2,
\end{equation}
where $\text{NN}(g)$ denotes the spatial neighbors of Gaussian $g$, 
typically selected within a fixed-radius region in the canonical space.  
This regularization encourages spatially adjacent Gaussians to share consistent motion coefficients 
and preserves locally smooth motion patterns across the canonical space.
\noindent Finally, the overall motion objective integrates both terms under the rigidity flag:
\begin{equation}
    \mathcal{L}_{\text{motion}} 
    =
    \sum_{k=1}^{K} 
    \Big[
    \motiongroup{\tau}{k} \, \mathcal{L}_{\text{rigid}}^{(k)}
    +
    (1-\motiongroup{\tau}{k})\, \mathcal{L}_{\text{nr}}^{(k)}
    \Big].
\end{equation}
Additional optimization objectives are described in the supplementary material.
With the object motion-aware optimization stage, we obtain a refined 4D Gaussian representation where rigid groups follow coherent $SE(3)$ motion and non-rigid groups exhibit locally smooth deformation. 
This reduces per-Gaussian drift and improves temporal consistency of the reconstructed motion.

\begin{table*}[ht!]
\centering
\caption{\textbf{Forecasting results on the iPhone dataset.}
We forecast frames beyond the observed time window and evaluate the predicted frames rendered from held-out test viewpoints.
The leftmost column (\emph{Obs. ratio}) indicates the fraction of input training frames used by each model; 80\% and 60\% correspond to forecasting the remaining 20\% and 40\% of frames, respectively.
$^{\dagger}$ denotes methods that are re-implemented for the SoM-based 4DGS representation and retain their original forecasting scheme.
}
\vspace{-2mm}
\resizebox{\textwidth}{!}{
\begin{tabular}{llcccccccccccc}
\toprule
\multirow{2}{*}{\emph{Obs. ratio}} & \multirow{2}{*}{Scene} 
& \multicolumn{3}{c}{GSPred~\cite{zhao2024gaussianprediction}} 
& \multicolumn{3}{c}{GSPred-SoM$^{\dagger}$~\cite{zhao2024gaussianprediction}} 
& \multicolumn{3}{c}{ODE-GS-SoM$^{\dagger}$~\cite{wang2025ode}}
& \multicolumn{3}{c}{{\oursname} (Ours)} \\
\cmidrule(lr){3-5} \cmidrule(lr){6-8} \cmidrule(lr){9-11} \cmidrule(lr){12-14}
& 
& mPSNR$\uparrow$ & mSSIM$\uparrow$ & mLPIPS$\downarrow$
& mPSNR$\uparrow$ & mSSIM$\uparrow$ & mLPIPS$\downarrow$
& mPSNR$\uparrow$ & mSSIM$\uparrow$ & mLPIPS$\downarrow$
& mPSNR$\uparrow$ & mSSIM$\uparrow$ & mLPIPS$\downarrow$ \\
\midrule
\multicolumn{1}{c}{\multirow{6}{*}{80\%}} 
& \multicolumn{1}{|l|}{apple}          & 15.75 & 0.5589 & \textbf{0.4338} & \multicolumn{1}{|c}{17.22} & 0.8122 & 0.4806 & \multicolumn{1}{|c}{16.35} & 0.8135 & 0.4803 & \multicolumn{1}{|c}{\textbf{17.97}} & \textbf{0.8199} & 0.4341 \\
& \multicolumn{1}{|l|}{block}          & 12.61 & 0.5220 & 0.5363 & \multicolumn{1}{|c}{12.20} & 0.5807 & \textbf{0.5352} & \multicolumn{1}{|c}{12.07} & 0.5717 & 0.5583 & \multicolumn{1}{|c}{\textbf{12.62}} & \textbf{0.5875} & \textbf{0.5352} \\
& \multicolumn{1}{|l|}{paper-windmill} & 14.76 & 0.2244 & 0.4329 & \multicolumn{1}{|c}{19.16} & \textbf{0.5573} & 0.2142 & \multicolumn{1}{|c}{19.13} & 0.5512 & 0.2154 & \multicolumn{1}{|c}{\textbf{19.34}} & 0.5552 & \textbf{0.2052} \\
& \multicolumn{1}{|l|}{spin}           & 14.54 & 0.4652 & 0.4002 & \multicolumn{1}{|c}{15.12} & 0.6701 & \textbf{0.3166} & \multicolumn{1}{|c}{15.69} & 0.6593 & 0.3354 & \multicolumn{1}{|c}{\textbf{15.96}} & \textbf{0.6740} & 0.3236 \\
& \multicolumn{1}{|l|}{teddy}          & 11.15 & 0.5790 & \textbf{0.5752} & \multicolumn{1}{|c}{11.26} & \textbf{0.5823} & 0.6942 & \multicolumn{1}{|c}{10.06} & 0.5821 & 0.7093 & \multicolumn{1}{|c}{\textbf{11.99}} & 0.5607 & 0.6157 \\
\rowcolor{avgrow}
\multicolumn{1}{c}{\cellcolor{white}} & \multicolumn{1}{|l|}{Average} & 13.76 & 0.4699 & 0.4757
& \multicolumn{1}{|c}{14.99} & \textbf{0.6405} & 0.4482
& \multicolumn{1}{|c}{14.66} & 0.6355 & 0.4597
& \multicolumn{1}{|c}{\textbf{15.58}} & 0.6395 & \textbf{0.4227} \\
\midrule
\multicolumn{1}{c}{\multirow{6}{*}{60\%}} 
& \multicolumn{1}{|l|}{apple}          & 14.44 & 0.5682 & 0.6108 & \multicolumn{1}{|c}{\textbf{16.96}} & \textbf{0.7460} & \textbf{0.4491} & \multicolumn{1}{|c}{15.48} & 0.7333 & 0.5035 & \multicolumn{1}{|c}{16.57} & 0.7409 & 0.4660 \\
& \multicolumn{1}{|l|}{block}          & 13.61 & 0.5494 & 0.5242 & \multicolumn{1}{|c}{13.02} & 0.6022 & 0.5231 & \multicolumn{1}{|c}{13.94} & 0.6101 & 0.5314 & \multicolumn{1}{|c}{\textbf{14.06}} & \textbf{0.6134} & \textbf{0.5055} \\
& \multicolumn{1}{|l|}{paper-windmill} & 14.96 & 0.2254 & 0.4651 & \multicolumn{1}{|c}{18.30} & \textbf{0.4888} & 0.2345 & \multicolumn{1}{|c}{17.94} & 0.4745 & 0.2566 & \multicolumn{1}{|c}{\textbf{18.34}} & 0.4858 & \textbf{0.2308} \\
& \multicolumn{1}{|l|}{spin}           & 14.52 & 0.4564 & 0.4203 & \multicolumn{1}{|c}{15.79} & 0.6719 & 0.3192 & \multicolumn{1}{|c}{14.89} & 0.6490 & 0.3508 & \multicolumn{1}{|c}{\textbf{16.47}} & \textbf{0.6838} & \textbf{0.3159} \\
& \multicolumn{1}{|l|}{teddy}          & 11.49 & \textbf{0.5828} & \textbf{0.5753} & \multicolumn{1}{|c}{11.63} & 0.5447 & 0.6020 & \multicolumn{1}{|c}{11.96} & 0.5526 & 0.6010 & \multicolumn{1}{|c}{\textbf{12.12}} & 0.5478 & 0.6045 \\
\rowcolor{avgrow}
\multicolumn{1}{c}{\cellcolor{white}} & \multicolumn{1}{|l|}{Average} & 13.80 & 0.4764 & 0.5192
& \multicolumn{1}{|c}{15.14} & 0.6107 & 0.4256
& \multicolumn{1}{|c}{14.84} & 0.6039 & 0.4487
& \multicolumn{1}{|c}{\textbf{15.51}} & \textbf{0.6143} & \textbf{0.4245} \\
\bottomrule
\end{tabular}
}
\label{tab:forecast_dycheck_view_iphone}
\vspace{-4mm}
\end{table*}

\subsection{Group-wise Motion Forecasting}
\label{sec:motion-forecasting}
After obtaining motion-consistent 4DGS representation, 
we aim to forecast the future evolution of Gaussian motion beyond the observed frames. 
At time $t$, the motion of a Gaussian $g \in \mathcal{G}$ is represented by its $SE(3)$ transformation 
$\mathbm{T}_{t,g} = [\deformed{R}{t,g} \mid \deformed{\mu}{t,g}]$.  
Similar to \cite{zhao2024gaussianprediction, wang2025ode}, 
our forecaster takes the optimized motion sequence from 4DGS representation, 
$\{\mathbm{T}_{t,g}\}_{t=0}^{T}$, 
as input and predicts its future continuation 
$\{\hat{\mathbm{T}}_{t,g}\}_{t > T}$.  
The forecasting is performed in an autoregressive rollout, iteratively using the latest $T\!-\!1$ frames including predictions to generate subsequent timesteps.

\vspace{-4mm}
\paragraph{Masked Input Training.}  
The motion forecaster adopts a shallow Transformer encoder~\cite{vaswani2017attention}, 
which effectively models temporal dependencies but tends to overfit when trained on fully observed sequences.  
To enhance generalization and temporal reasoning, 
we introduce a masked motion modeling strategy inspired by masked language modeling in NLP~\cite{devlin2019bert}.  
This segment-level masking encourages the model to infer missing temporal dynamics 
from surrounding motion context, resulting in better temporal understanding.  
It further improves long-horizon prediction and robustness to noisy inputs.  
The masking ratio is gradually annealed during training to match the inference condition.

\vspace{-3mm}
\paragraph{Group-wise Forecasting.}
We train a lightweight motion forecaster to each motion group $\motiongroup{G}{k}$ obtained from the group-wise optimization in Section~\ref{sec:group-optimization}, enabling the model to learn group-specific temporal dynamics.
This group-wise formulation enhances both stability and accuracy of motion forecasting by decoupling heterogeneous motion dynamics across objects.
Within each group, the forecaster captures consistent temporal patterns shared among Gaussians exhibiting similar motion behavior, leading to smoother and more coherent predictions.

\vspace{-3mm}
\paragraph{Training Objectives.}
For each motion group $\motiongroup{G}{k}$, the forecaster takes the Gaussian motion up to time $T\!-\!1$ as input and is trained to minimize the discrepancy between the predicted and observed motion at time $T$.  
The motion reconstruction loss is defined as
\begin{equation}
\mathcal{L}_{\text{pred}}^{(k)}
= \frac{1}{|\motiongroup{G}{k}|}
\sum_{g \in \motiongroup{G}{k}}
\big\|
\canonicalsub{T}{T,g} - \canonicalsub{\hat{T}}{T,g}
\big\|_2^2.
\end{equation}
To enforce physically smooth motion, we apply an acceleration regularization term:
\begin{equation}
\mathcal{L}_{\text{acc}}^{(k)}
= \frac{1}{|\motiongroup{G}{k}|}
\sum_{g \in \motiongroup{G}{k}}
\big\|
\deformed{\hat{\mu}}{T,g}
- 2\deformed{{\mu}}{T-1,g}
+ \deformed{{\mu}}{T-2,g}
\big\|_2^2,
\end{equation}
where $\deformed{\hat{\mu}}{t,g}$ denotes the predicted mean of Gaussian $g$ at time $t$.
The overall training objective for $\motiongroup{G}{k}$ is given by:
\begin{equation}
\mathcal{L}_{\text{group}}^{(k)} 
= \mathcal{L}_{\text{pred}}^{(k)} 
+ \lambda_{\text{acc}}\, \mathcal{L}_{\text{acc}}^{(k)},
\end{equation}
and each forecaster is trained independently using $\mathcal{L}_{\text{group}}^{(k)}$.

%% file: sec/4_experiments.tex
\section{Experiments}
\subsection{Implementation Details}
We build our dynamic reconstruction pipeline upon Shape-of-Motion (SoM)~\cite{wang2025shape}, which serves as the backbone for canonical space and motion parameterization. For Gaussian grouping, we extend the official implementation of Gaga~\cite{lyu2024gaga} and integrate it into our framework.
For motion forecasting, we design a lightweight transformer-based forecaster consisting of a single-layer encoder with eight attention heads, a 32-dimensional feature embedding, and a 64-dimensional feed-forward expansion. A final prediction head maps the encoded features to future motion parameters.
Additional architectural and optimization details are provided in the Supplementary Material (\Cref{sup_sec:impl_detail}).

\begin{figure*}[ht]
\centering
\includegraphics[width=\textwidth]{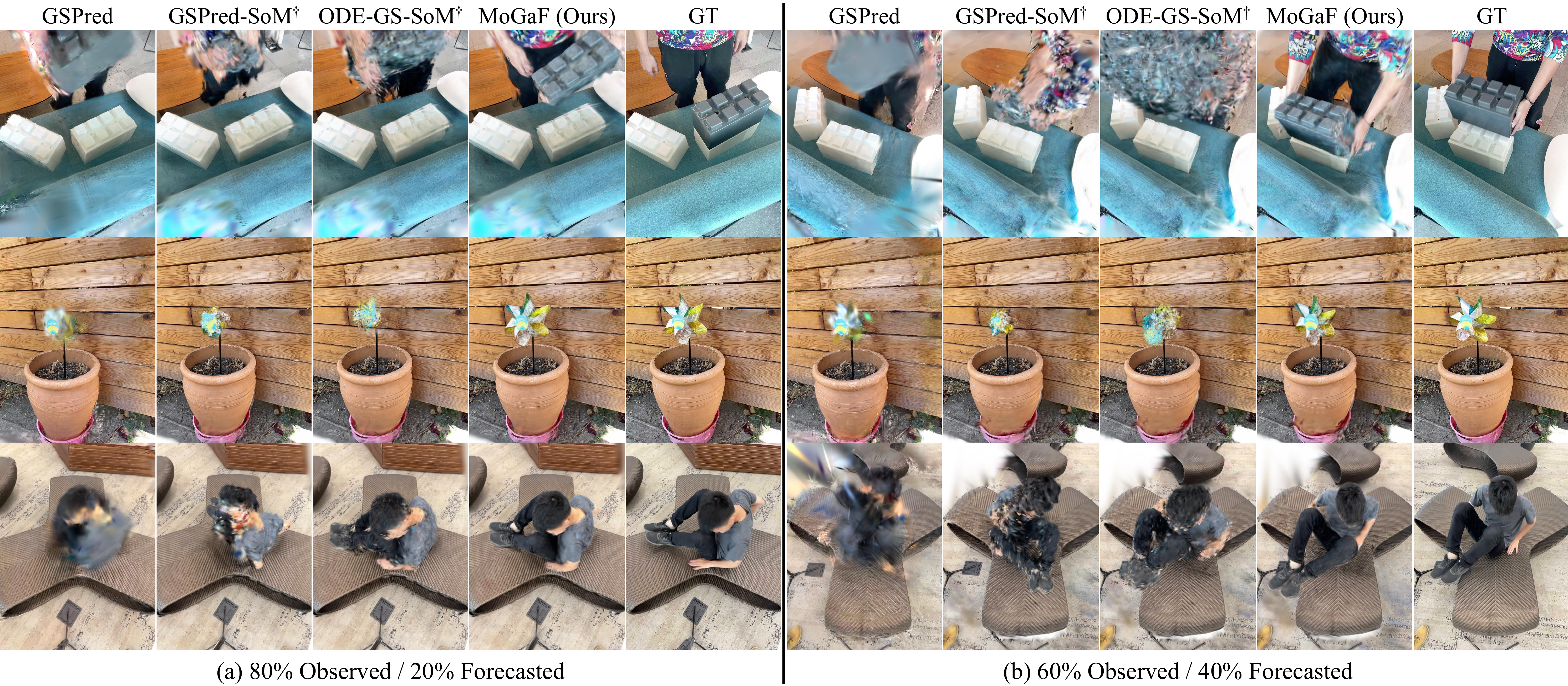}
\vspace{-7mm}
\caption{\textbf{Qualitative results on iPhone dataset.}
We present forecasted frames from test camera views. 
(a) and (b) correspond to settings where the first 80\% and 60\% of frames are used for training, and the remaining 20\% and 40\% are forecasted, respectively.}
\label{fig:main_iphone_qual}
\vspace{-6mm}
\end{figure*}

\subsection{Future View Synthesis}
We evaluate future view synthesis performance following the protocol of GSPred~\cite{zhao2024gaussianprediction}. 
Specifically, we use a subset of the original training frames as input observations and measure forecasting performance on the remaining unseen frames. 
We consider two observation ratios: $80\%$ (as in GSPred) and a more challenging $60\%$ setting to assess long-term prediction capability.

\vspace{-3mm}
\paragraph{Evaluation on iPhone dataset.}
We first evaluate our method on iPhone dataset~\cite{gao2022monocular}, 
which contains hundreds of frames of dynamic scenes captured from multiple handheld cameras.  
Since GSPred~\cite{zhao2024gaussianprediction} is built on \cite{wu20244d}, 
it does not leverage data-driven priors such as depth estimation~\cite{yang2024depth} 
or dense point tracking~\cite{karaev2024cotracker, doersch2023tapir}, 
and thus often underperforms on highly dynamic monocular captures.  
For a fair comparison, we adopt the GCN-based forecasting architecture from GSPred and train it on top of SoM, denoted as \textit{GSPred-SoM}.
Additionally, following~\cite{wang2025ode}, we reproduce a latent neural ODE forecaster on the SoM representation, denoted as \textit{ODE-GS-SoM}.

\begin{figure*}[ht!]
\centering
\includegraphics[width=\textwidth]{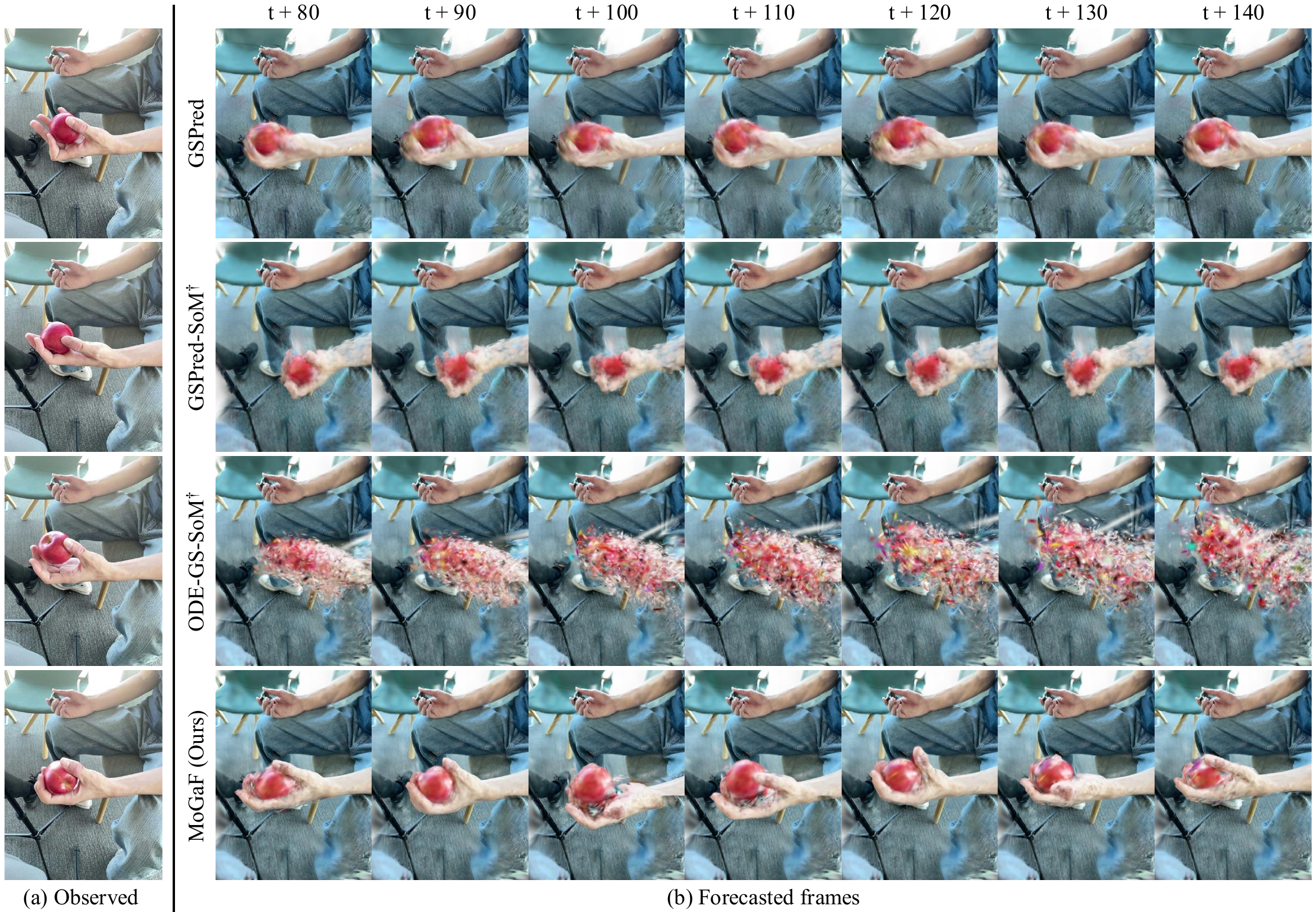}
\vspace{-6mm}
\caption{\textbf{Long-term forecasting results on iPhone dataset.}
(a) shows GT views from test cameras at the observed timesteps, and (b) presents the forecasted renderings for timesteps beyond the observations. Note that $t$ denotes the timestep of the last observed frame.}
\label{fig:iphone_long_term_forecast}
\vspace{-6mm}
\end{figure*}

As shown in Table~\ref{tab:forecast_dycheck_view_iphone}, \oursname\ consistently achieves higher photometric fidelity than the baselines.
Figure~\ref{fig:main_iphone_qual} shows that \oursname\ preserves the geometry of both rigid and non-rigid dynamic objects, producing realistic future renderings by accurately extrapolating the observed motion. In contrast, baselines often exhibit geometric inconsistency and unrealistic dynamics in scenes involving large or rapid motion.
To further assess long-term behavior, we examine the forecasting results for \emph{apple} in Figure~\ref{fig:iphone_long_term_forecast}. 
GSPred renders the hand as nearly static in the future frames despite its highly dynamic motion in the observations, while GSPred-SoM and ODE-GS-SoM fail to maintain object geometry. 
In comparison, \oursname\ accurately preserves object geometry and predicts the hand’s complex motion, demonstrating its effectiveness in extrapolating coherent dynamics.
Additional results are included in the Supplementary Material.

\vspace{-2mm}
\begin{table}[t!]
\centering
\caption{\textbf{Forecasting results on D-NeRF dataset.} We use 60\% of the frames as training and predict the remaining 40\%.}
\vspace{-2mm}
\resizebox{\columnwidth}{!}{
\begin{tabular}{lcccccc}
\toprule
\multirow{2}{*}{Scene} & \multicolumn{3}{c}{GSPred~\cite{zhao2024gaussianprediction}} & \multicolumn{3}{c}{Ours} \\
\cmidrule(lr){2-4} \cmidrule(lr){5-7}
 & PSNR$\uparrow$ & SSIM$\uparrow$ & LPIPS$\downarrow$
 & PSNR$\uparrow$ & SSIM$\uparrow$ & LPIPS$\downarrow$ \\
\midrule
\multicolumn{1}{l|}{Trex}            & 20.63 & 0.9271 & \multicolumn{1}{c|}{0.0627} & \textbf{20.80} & \textbf{0.9379} & \textbf{0.0551} \\
\multicolumn{1}{l|}{Mutant}          & 25.14 & \textbf{0.9375} & \multicolumn{1}{c|}{\textbf{0.0415}} & \textbf{25.69} & 0.9246 & 0.0482 \\
\multicolumn{1}{l|}{Jumpingjacks}    & 19.84 & 0.9087 & \multicolumn{1}{c|}{0.0924} & \textbf{19.86} & \textbf{0.9090} & \textbf{0.0920} \\
\multicolumn{1}{l|}{Standup}         & 21.99 & 0.9170 & \multicolumn{1}{c|}{0.0796} & \textbf{22.32} & \textbf{0.9201} & \textbf{0.0756} \\
\multicolumn{1}{l|}{Bouncingballs}   & 23.68 & 0.9603 & \multicolumn{1}{c|}{0.0540} & \textbf{25.03} & \textbf{0.9623} & \textbf{0.0436} \\
\multicolumn{1}{l|}{Hook}            & 21.18 & 0.8855 & \multicolumn{1}{c|}{0.0851} & \textbf{22.41} & \textbf{0.8929} & \textbf{0.0784} \\
\multicolumn{1}{l|}{Hellwarrior}     & 29.11 & 0.9056 & \multicolumn{1}{c|}{0.0957} & \textbf{29.23} & \textbf{0.9075} & \textbf{0.0894} \\
\multicolumn{1}{l|}{Lego}            & 12.65 & 0.7667 & \multicolumn{1}{c|}{0.2244} & \textbf{21.61} & \textbf{0.8631} & \textbf{0.1142} \\
\midrule
\multicolumn{1}{l|}{\textbf{Average}}& 21.78 & 0.9011 & \multicolumn{1}{c|}{0.0919} & \textbf{23.37} & \textbf{0.9147} & \textbf{0.0746} \\
\bottomrule
\end{tabular}
}
\vspace{-6mm}
\label{tab:dnerf_0.6}
\end{table}

\begin{figure}[ht!]
\centering
\includegraphics[width=\columnwidth]{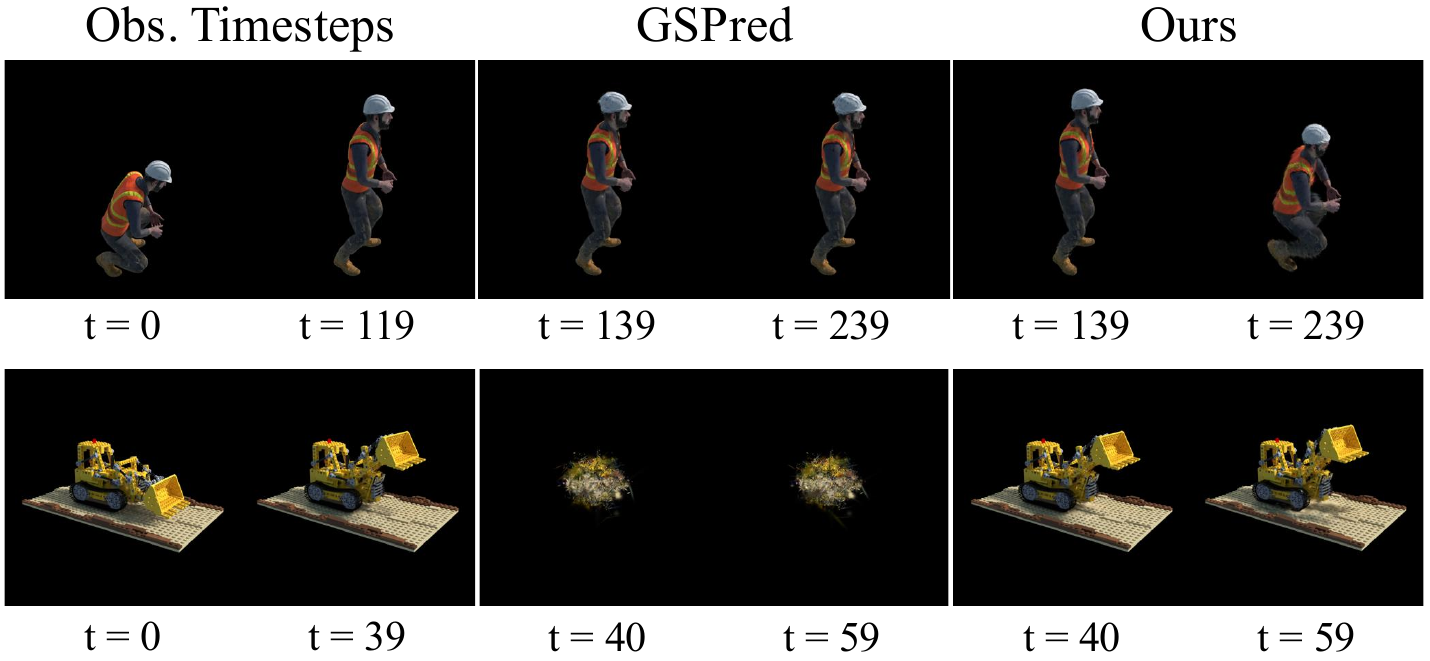}
\vspace{-6mm}
\caption{\textbf{Forecasting results on D-NeRF dataset.}
We render extrapolated future frames. Note that in \emph{Obs. Timesteps}, the first and second columns show renderings reconstructed from training views at the first and last observed timesteps, respectively.}
\label{fig:dnerf_0.8_forecast}
\vspace{-6mm}
\end{figure}

\vspace{-3mm}
\paragraph{Evaluation on D-NeRF dataset.}
We also evaluate forecasting performance on D-NeRF dataset~\cite{pumarola2021d}, which consists of synthetic frames capturing dynamic objects rendered from randomly sampled viewpoints on a hemisphere at each timestep.
Since D-NeRF dataset consists solely of object-level scenes, we follow the same 4DGS training procedure as GSPred~\cite{zhao2024gaussianprediction} to enable a fair comparison of the forecasting module in \oursname.
As shown in Table~\ref{tab:dnerf_0.6}, {\oursname} achieves higher performance across most
scenes, demonstrating stronger temporal coherence and long-range motion modeling under the 
60\%–40\% forecasting setting.  
Figure~\ref{fig:dnerf_0.8_forecast} further presents qualitative results under the 80\%–20\% split.  
In \emph{Standup}, \oursname\ captures the global motion trend and produces realistic long-term forecasts. In \emph{Lego}, it preserves object structure and motion, whereas GSPred often loses shape consistency. Additional results are provided in the Supplementary Material.

\subsection{Ablation Studies}

\paragraph{Group-wise Optimization and Forecasting.}

We analyze the effectiveness of our group-wise optimization and forecasting. To this end, we remove the group-wise optimization stages and forecast the future motion of all dynamic Gaussians jointly with a single forecaster (i.e., without object-wise forecasters).
We evaluate forecasting quality using 2D and 3D point-tracking metrics following the protocol of \cite{wang2025shape}.
As shown in Table~\ref{tab:ablation_tracking}, performance consistently degrades without the group-aware design. These results indicate that object-level structure is crucial for physically plausible, temporally coherent long-term forecasts.
More detailed ablations of the group-wise modules are reported in the Supplementary Material~\ref{sup_sec:group-optim-analysis} (Table~\ref{sup_tab:ablation_groupwise}).

\begin{figure}[t]
\centering
\includegraphics[width=0.9\columnwidth]{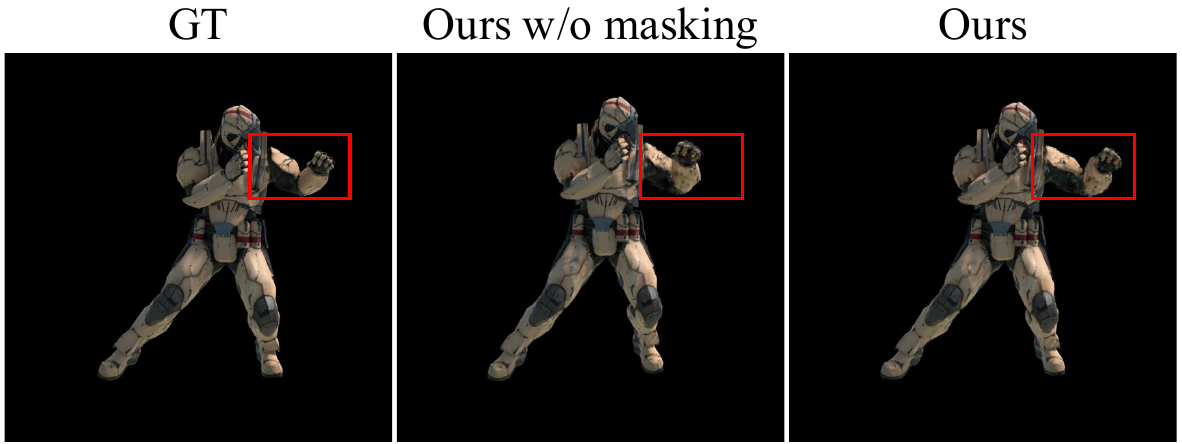}
\vspace{-2mm}
\caption{\textbf{Effect of masking.} 
Applying contiguous-span masking during training enhances the robustness of motion forecasting.}
\label{fig:ablation_dnerf_masking}
\vspace{-6mm}
\end{figure}

\vspace{-4mm}
\paragraph{Masked Motion Forecaster Training.}
We assess the effect of our motion-masking strategy on D-NeRF dataset by comparing the forecaster against a variant trained without masking.
As shown in Table~\ref{tab:ablation_dnerf_masking} and Figure~\ref{fig:ablation_dnerf_masking}, removing the masking strategy leads to noticeable performance degradation.
This gap indicates that masking helps the encoder attend to intrinsic motion cues, resulting in more stable long-term predictions and improved robustness under extended forecasting horizons.

\vspace{-2mm}
\begin{table}[!t]
\centering
\caption{\textbf{Ablation results.} 
(Top) Tracking performance on iPhone dataset. 
(Bottom) Masking performance on D-NeRF dataset.}
\vspace{-2mm}
\begin{subtable}{\linewidth}
\centering
\caption{
We evaluate point-tracking metrics on the forecasted 40\% future frames with and without both group-wise optimization and forecasting.
}
\resizebox{\linewidth}{!}{
    \begin{tabular}{lcccccc}
        \toprule
        \multirow{2}{*}{Method} & \multicolumn{3}{c}{3D tracking} & \multicolumn{3}{c}{2D tracking} \\
        \cmidrule(lr){2-4} \cmidrule(lr){5-7}
         & EPE$\downarrow$ & $\delta_{3D}^{.10}$$\uparrow$ & $\delta_{3D}^{.05}$$\uparrow$
         & AJ$\uparrow$ & $\delta_{avg}$$\uparrow$ & OA$\uparrow$ \\
        \midrule
        Ours w/o grouping
        & 0.296 & 35.6 & 17.1 & 13.3 & 8.5 & 64.1 \\
        {\oursname} (Ours)
        & \textbf{0.236} & \textbf{44.8} & \textbf{22.5} & \textbf{17.5} & \textbf{8.8} & \textbf{80.1} \\
        \bottomrule
    \end{tabular}
}
\label{tab:ablation_tracking}
\end{subtable}

\vspace{1mm}
\begin{subtable}{0.7\linewidth}
\centering
\caption{
We evaluate 20\% future frame forecasting using forecasters trained with and without masking.}

\resizebox{\linewidth}{!}{
\begin{tabular}{l c c c}
\toprule
Method & PSNR$\uparrow$ & SSIM$\uparrow$ & LPIPS$\downarrow$ \\
\midrule
Ours w/o masking & 24.68 & 0.9283 & 0.0551 \\
Ours             & \textbf{25.87} & \textbf{0.9357} & \textbf{0.0491} \\
\bottomrule
\end{tabular}
}
\label{tab:ablation_dnerf_masking}
\end{subtable}
\vspace{-6mm}
\end{table}

%% file: sec/5_conclusion.tex
\section{Conclusion}
We introduced Motion Group-aware Gaussian Forecasting (\oursname), a unified framework for dynamic scene long-term extrapolation leveraging structured 4D representation. {\oursname} explicitly models object-level dynamics through motion-aware Gaussian grouping, group-wise optimization, and a lightweight forecasting module. By enforcing physically consistent rigid and non-rigid motion within a structured space-time representation, {\oursname} produces 4DGS models that are both spatially coherent and temporally stable, enabling realistic scene evolution beyond the observed frames. Extensive experiments on synthetic and real-world datasets show that {\oursname} significantly improves long-term forecasting accuracy over existing baselines, and ablation studies confirm the contribution of each component to achieving spatial consistency and stable predictions.

%% file: sec/6_acknowledgement.tex
\section*{Acknowledgement}
\label{sec:acknowledgement}
\small
This work was supported by the IITP grants (RS-2022-II220959: Few-Shot Learning of Causal Inference in Vision and Language for Decision Making (50\%), RS-2022-II220264: Comprehensive Video Understanding and Generation (25\%), RS-2022-II220290: Visual Intelligence for Space-Time Understanding and Generation based on Multi-layered Visual Common Sense (20\%), RS-2019-II191906: AI Graduate School at POSTECH (5\%)) funded by the Ministry of Science and ICT, Korea. 

%% file: supple_sec/2_grouping.tex
\section{Details of Gaussian Grouping}
\label{sup_sec:grouping_details}
In this section, we provide the detailed algorithms used in our Gaussian grouping pipeline, complementing the description in \Cref{sec:gs-grouping} of the main paper.  
We first summarize the static Gaussian grouping mechanism used in Gaga~\cite{lyu2024gaga} (\Cref{sup_sec:gaga_static}),  
then describe its naive 4D extension (\Cref{sup_sec:gaga4D}),  
and finally introduce our motion-aware grouping algorithm that incorporates spatiotemporal cues (\Cref{sup_sec:ours_grouping}). 

\subsection{Preliminary: Static Gaussian Grouping}
\label{sup_sec:gaga_static}

Gaga~\cite{lyu2024gaga} formulates static 3D Gaussian grouping as the problem of assigning each 3D Gaussian to group label using multi-view 2D instance masks.  
Given a set of static 3D Gaussians $\mathcal{G}$ reconstructed from a scene and multi-view images with segmentation masks, 
each 2D mask $M$ is first associated with Gaussians those projections fall inside the mask region:
\begin{equation}
    \mathcal{G}(\motiongroup{M}{k}_i)
    =
    \big\{
        g \in \mathcal{G}
        \;\big|\;
        \mathrm{Proj}(g) \in \motiongroup{M}{k}_i
    \big\},
\end{equation}
where $\motiongroup{M}{k}_i$ is k-th mask detected in $i$-th image.

\paragraph{3D-aware memory bank.}
To enforce cross-view consistency, Gaga employs memory banks storing Gaussian groups $\{\motiongroup{\mathcal{G}}{k}\}$,  
where each group represents an object-level region in 3D space.  
At initialization, the masks from the first view are each inserted into the memory bank as separate groups 
by setting $\motiongroup{\mathcal{G}}{k} \leftarrow \mathcal{G}(\motiongroup{M}{k}_0)$ for every mask $\motiongroup{M}{k}_0$ in the initial view.  
For subsequent views, new masks are either merged into one of these existing groups or used to create additional groups,  
depending on their overlap with the current memory contents.

\begin{figure*}[ht!]
\centering
\includegraphics[width=0.9\textwidth]{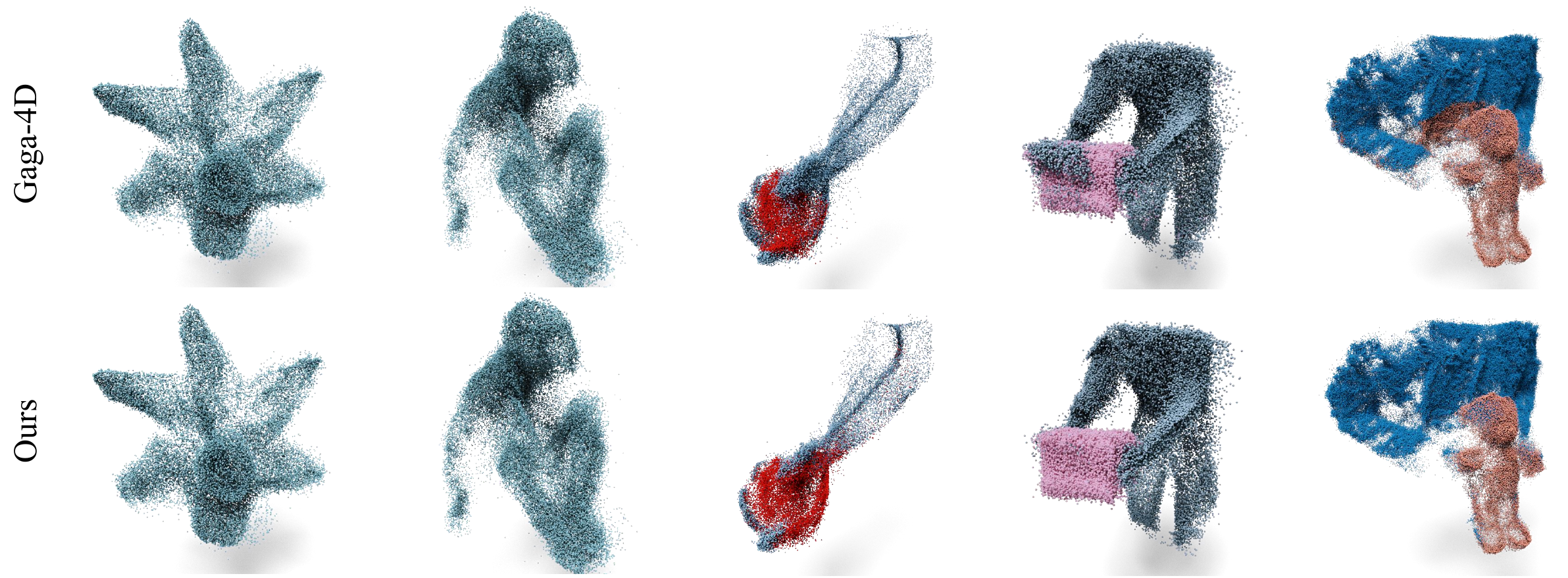}
\caption{\textbf{Gaussian grouping results.}
We present Gaussian grouping results of our method and Gaga-4D on iPhone dataset~\cite{gao2022monocular}, trained on the full sequence.}
\label{sup_fig:grouping_comparison}
\vspace{-4mm}
\end{figure*}

\paragraph{Group ID assignment.}
For a new mask $m = \motiongroup{M}{j}_i$ extracted from $i$-th input image, the similarity between its associated Gaussians $\mathcal{G}(m)$ and an existing group $\motiongroup{G}{k}$ 
is computed using the shared-Gaussian overlap:
\begin{equation}
    \mathrm{Overlap}(m, k)
    =
    \frac{
        \#\big( \motiongroup{\mathcal{G}}{k} \cap \mathcal{G}(m) \big)
    }{
        \#\big( \mathcal{G}(m) \big)
    }.
\end{equation}
This formulation depends only on the newly observed set $\mathcal{G}(m)$,  
so the threshold is independent of how many Gaussians have already been accumulated in $\motiongroup{\mathcal{G}}{k}$,  
avoiding the need to continually retune it as the memory bank grows.

Let $k^\star = \arg \max\limits_{k} \mathrm{Overlap}(m, k)$ be the group with the highest overlap score.  
If $\mathrm{Overlap}(m,k^\star)$ exceeds a fixed threshold $\theta_{\mathrm{ov}}$, Gaussians in $\mathcal{G}(m)$ are merged into group $\motiongroup{\mathcal{G}}{k^\star}$:
\begin{equation}
    \motiongroup{\mathcal{G}}{k^\star}
    \leftarrow
    \motiongroup{\mathcal{G}}{k^\star} \cup \mathcal{G}(m).
\end{equation}
Otherwise, a new group is created in the memory bank with $\mathcal{G}(m)$ as its initial Gaussian set.  
In practice, each Gaussian is assigned to at most one group by tracking canonical indices,  
so once a Gaussian has been inserted into a group, it is not duplicated elsewhere.  

\paragraph{Static setting.}
Because Gaga operates on static scenes, each Gaussian maintains a fixed 3D position,  
making 2D projection overlap a reliable cue for multi-view object grouping.  
The method produces coherent object-part segmentation and provides the foundation 
for extending grouping to dynamic scenes in our work.

\begin{algorithm}[t]
\caption{\textit{Naive 4D Gaussian Grouping}}
\label{alg:naive4D_compact}
\KwIn{Frames $\{I_t\}$, segmentation model $\mathcal{S}$, Gaussians $\mathcal{G}$, deformed states $\{g_t\}$, overlap threshold $\theta_{\mathrm{ov}}$}
\KwOut{Groups $\{\motiongroup{\mathcal{M}}{k}=(\motiongroup{G}{k},\motiongroup{\tau}{k})\}$}

\textbf{Init (first keyframe):}
Obtain $\{M^{(k)}_{t_0},\tau^{(k)}_{\mathrm{mask}}\}=\mathcal{S}(I_{t_0})$;
\For{$k=1..K$}{
  $\motiongroup{G}{k}\leftarrow\{g\in\mathcal{G}\mid\mathrm{Proj}(g_{t_0})\in M^{(k)}_{t_0}\}$;
  $\motiongroup{\tau}{k}\leftarrow\tau^{(k)}_{\mathrm{mask}}$;
}
Define canonical ID $\mathrm{id}(g)$ for all $g$.

\textbf{Iterative grouping across frames:}
\For{$t=1..T$}{
  Obtain $\{M^{(k)}_t,\tau^{(k)}_{\mathrm{mask}}\}$;\\
  \For{$k=1..K$}{
    $\motiongroup{G}{k}_t\leftarrow\{g\in\mathcal{G}\mid\mathrm{Proj}(g_t)\in M^{(k)}_t\}$;
    Compute $\mathrm{Overlap}(t,k)$ for all $k$ using canonical IDs;\\
    \If{$\mathrm{Overlap}(t,k)\ge\theta_{\mathrm{ov}}$}{
       $\motiongroup{G}{k}\leftarrow \motiongroup{G}{k}\cup\motiongroup{G}{k}_t$;
    }
  }
}
\end{algorithm}

\subsection{Naive Extension of Static Gaussian Grouping}
\label{sup_sec:gaga4D}

\paragraph{4D-aware Memory Bank.}
Given a 2D mask $\motiongroup{M}{k}_t$ at timestep $t$, we extract deformed Gaussians contributing to the mask:
\begin{equation}
\label{eq:4d-proj}
    \motiongroup{G}{k}_t
    =
    \big\{
        g \in \mathcal{G}
        \;\big|\;
        \mathrm{Proj}(g_t) \in \motiongroup{M}{k}_t
    \big\},
\end{equation}
where $g_t$ denotes the deformed state of $g$.  
We extend the static memory bank to maintain motion groups over time.  
At the first frame, the $k$-th memory bank is initialized as
\begin{equation}
    \motiongroup{\mathcal{M}}{k}
    =
    \left(\motiongroup{G}{k}_t,\; \motiongroup{\tau}{k}_{\mathrm{mask}}\right),
\end{equation}
where $\motiongroup{\tau}{k}_{\mathrm{mask}}$ denotes the rigidity type of the segmented region.

\paragraph{Group ID Assignment.}
Different from Gaga~\cite{lyu2024gaga}, we have access to consistent mask identities provided by video segmentation models~\cite{ren2024grounding, ravi2024sam},  
which serve as a temporal cue for associating $\motiongroup{G}{k}_t$ with the same object across frames.  
Therefore, we do not apply the max-IoU–based merging strategy used for label-agnostic matching in static Gaussian segmentation.
However, dynamic deformation and occlusions may produce unreliable matches, so we verify consistency using the shared-Gaussian overlap:
\begin{equation}
    \mathrm{Overlap}(t,k)
    =
    \frac{
        \#\!\left( \motiongroup{G}{k} \cap \motiongroup{G}{k}_t \right)
    }{
        \#\!\left( \motiongroup{G}{k}_t \right)
    }.
\end{equation}
The intersection is defined through canonical identity:
\begin{equation}
\small{
    \motiongroup{G}{k} \cap \motiongroup{G}{k}_t
    =
    \left\{
        g \in \motiongroup{G}{k}_t
        \;\middle|\;
        \exists\, g' \in \motiongroup{G}{k},\;
        \mathrm{id}(g)=\mathrm{id}(g')
    \right\}.
}
\end{equation}

\noindent
If $\mathrm{Overlap}(t,k)\ge\theta_{\mathrm{ov}}$, we merge:
\begin{equation}
    \motiongroup{G}{k}
    \leftarrow
    \motiongroup{G}{k} \cup \motiongroup{G}{k}_t.
\end{equation}

\paragraph{Limitation of naive 4D extension.}
Because this naive extension relies solely on instantaneous 2D projection overlap,  
Gaussian deformation from motion optimization (e.g., occlusions, drift, partial visibility) can cause inconsistent overlap patterns.  
As a result, Gaussians belonging to the same object may fail to merge, fragmenting a single object into multiple groups,  
while unrelated objects may be mistakenly merged due to projected overlap.  
This makes the naive extension fragile under ambiguous projections (Figure~\ref{fig:grouping_comparison}a of the main paper),  
highlighting the need for motion-aware grouping that leverages temporal and motion cues beyond simple projection overlap. 

\subsection{Motion-aware Gaussian Grouping}
\label{sup_sec:ours_grouping}
\begin{algorithm}[t!]
\caption{\textit{Complete Procedure of Motion-aware Gaussian Grouping}}
\label{alg:gaussian_grouping_complete}
\KwIn{Video frames $\{I_t\}_{t=1}^T$, segmentation model $\mathcal{S}$, dynamic Gaussian set $\mathcal{G}$ with deformed states $\{ g_t \}_{t=1}^T$}
\KwOut{Motion groups $\{\motiongroup{\mathcal{M}}{k}=(\motiongroup{G}{k},\,\motiongroup{\tau}{k})\}_{k=1}^K$}

\BlankLine
\textbf{Stage 1: Init at the First Keyframe}

Obtain $\{\motiongroup{M}{k}_t,\,\motiongroup{\tau}{k}_{\text{mask}}\} \leftarrow \mathcal{S}(\{I_t\})$ \\
Select first keyframe $t_k$ \\
\ForEach{$k=1,\dots,K$}{
  $\motiongroup{G}{k} \leftarrow \{\, g\in\mathcal{G} \mid \text{Proj}(g_{t_k}) \in \motiongroup{M}{k}_{t_k}\}$ \\
  $\motiongroup{\tau}{k} \leftarrow \motiongroup{\tau}{k}_{\text{mask}}$
}
Define features for all $g$: $\bm{f}_g = [\,\canonicalsub{\mu}{c,g},\,\bm{w}'_g\,]$

\BlankLine
\textbf{Stage 2: Iterative Region Growing}
\ForEach{keyframe $t$ in stride order}{

  \tcp{(A) Feature-space region growing}
  \Repeat{\textit{no group changes}}{
    $U \leftarrow \mathcal{G} \setminus \bigcup_k \motiongroup{G}{k}$ \\
    \ForEach{$k=1,\dots,K$}{
      $\epsilon_k \leftarrow \alpha \cdot \mathrm{mean}_{g\in\motiongroup{G}{k}}\!\left[\mathrm{KNN}_K(\bm{f}_g)\right]$ \\
      $N_k \leftarrow \{\, g' \in U \mid \min_{g\in\motiongroup{G}{k}}\|\bm{f}_g - \bm{f}_{g'}\| < \epsilon_k \,\}$ \\
      $\motiongroup{G}{k} \leftarrow \motiongroup{G}{k} \cup N_k$ \\
      $U \leftarrow U \setminus N_k$
    }
  }

  \tcp{(B) Seeding \& merge at keyframe $t$}
  \ForEach{$k=1,\dots,K$}{
    $\motiongroup{G}{k}_t \leftarrow \{\, g \mid \text{Proj}(g_t) \in \motiongroup{M}{k}_t \}$ \\
    $\motiongroup{G}{k} \leftarrow \motiongroup{G}{k} \cup \motiongroup{G}{k}_t$
  }
}

\BlankLine
\textbf{Stage 3: Reassign Unassigned Gaussians (KNN Voting)}

$G_{\text{labeled}} \leftarrow \bigcup_k \motiongroup{G}{k}$, 
$U \leftarrow \mathcal{G} \setminus G_{\text{labeled}}$ \\

\ForEach{$g_u \in U$}{
  $\mathcal{N}(g_u) \leftarrow \mathrm{KNN}_K(\bm{f}_{g_u},\, G_{\text{labeled}})$ \\[2pt]
  
  \ForEach{$k = 1,\dots,K$}{
    $v_k(g_u) \leftarrow 
      \sum_{g' \in \mathcal{N}(g_u)} 
      \mathbb{I}[\, g' \in \motiongroup{G}{k} \,]$
  }
  
  $k^\star(g_u) = \arg\max_k v_k(g_u)$ \\
  $\motiongroup{G}{k^\star(g_u)} 
      \leftarrow \motiongroup{G}{k^\star(g_u)} \cup \{ g_u \}$
}
\end{algorithm}
To address the limitations of the naive 4D extension, we propose a spatiotemporal Gaussian grouping algorithm tailored for pre-optimized 4DGS representations.  
Our method follows an alternating strategy composed of (i) keyframe-based seeding and (ii) feature-space region growing.  
We first sample $T_{\text{key}}$ keyframes from the input video,  
forming the keyframe set $\{I_t\}_{t=1}^{T_{\text{key}}}$,  
and extract Gaussians that contribute to the mask rendering at these frames.

\paragraph{Iterative Region Growing Strategy.}
We initialize motion groups  
$\motiongroup{\mathcal{M}}{k} = \{(\motiongroup{G}{k}, \motiongroup{\tau}{k})\}$  
by extracting the front Gaussians at a set of keyframes following \Cref{eq:4d-proj}.  
After initialization, we propagate group assignments to neighboring Gaussians via a spatiotemporal region-growing process.

Each Gaussian $g \in \mathcal{G}$ is represented by a spatiotemporal feature vector
\begin{equation}
    \bm{f}_g =
    \big[
        \deformed{\mu}{c,g},
        \;\deformed{w'}{g}
    \big],
\end{equation}
where $\deformed{\mu}{c,g}$ is the canonical-space mean of $g$, and  
$\deformed{w'}{g}$ denotes its PCA-reduced motion coefficient.

Given these features, each group $\motiongroup{G}{k}$ is expanded by aggregating Gaussians in its feature-space neighborhood.  
For every $g \in \motiongroup{G}{k}$, we consider its nearest neighbors and include any candidate $g'$ if
\begin{equation}
    \|\bm{f}_g - \bm{f}_{g'}\| < \epsilon_r,
    \qquad
    g' \notin \motiongroup{G}{k}.
\end{equation}
This update is repeated until convergence.  
We then perform a new seeding step at the next keyframe and apply region growing again.  
By alternating between these two stages, the grouping progressively stabilizes and captures coherent spatiotemporal structure across the sequence.

\paragraph{Handling Unassigned Gaussians.}
After the alternating seeding–region-growing stages, a small number of Gaussians may remain unassigned because
they were not included in any feature-space neighborhood expansion.
To relabel these remaining Gaussians, we perform a feature-space voting procedure using only group-labeled Gaussians.

For each unassigned Gaussian $g_u$, we retrieve its $K$ nearest labeled neighbors in feature space:
\begin{equation}
    \mathcal{N}(g_u)
    =
    \operatorname{KNN}_K(\bm{f}_{g_u},\, \mathcal{G}_{\text{labeled}}),
\end{equation}
where $\bm{f}_{g_u}$ is the spatiotemporal feature of $g_u$ and
$\mathcal{G}_{\text{labeled}} = \bigcup_k \motiongroup{G}{k}$ denotes the set of Gaussians with assigned group labels.
Among these labeled neighbors, we count how many belong to each motion group.  
Let $\mathbb{I}[g' \in \motiongroup{G}{k}]$ denote an indicator function that evaluates to 1 if $g'$ belongs to group $k$, and 0 otherwise.  
The voting score for group $k$ is computed as
\begin{equation}
    v_k(g_u)
    =
    \sum_{g' \in \mathcal{N}(g_u)}
    \mathbb{I}[g' \in \motiongroup{G}{k}].
\end{equation}
The final group assignment for $g_u$ is then obtained by majority voting:
\begin{equation}
    k^\star(g_u)
    =
    \arg\max_k\, v_k(g_u).
\end{equation}
We then update the motion group as
\begin{equation}
    \motiongroup{G}{k^\star}
    \leftarrow
    \motiongroup{G}{k^\star} \cup \{g_u\}.
\end{equation}
This voting-based reassignment ensures that isolated or boundary Gaussians are consistently grouped
according to their local spatiotemporal context.

As shown in \Cref{sup_fig:grouping_comparison}, our grouping algorithm works robustly across multiple objects.  
It produces clear and stable group assignments even in scenes where distinct objects exhibit significant spatial overlap in the input video.

%% file: supple_sec/1_implementation.tex
\section{Implementation Details}
\label{sup_sec:impl_detail}
\subsection{4DGS Initialization via SoM}
\label{sup_sec:pre-optim}
Our method builds on pre-optimized 4DGS representation following Shape-of-Motion (SoM)~\cite{wang2025shape}, 
trained for 300 epochs. We use the same optimization hyperparameters for all benchmark datasets, following the official SoM implementation. For the future view synthesis experiments, we use only the first 60\% or 80\% of the input sequence to evaluate extrapolation performance.

\begin{figure*}[t!]
\centering
\includegraphics[width=0.8\textwidth]{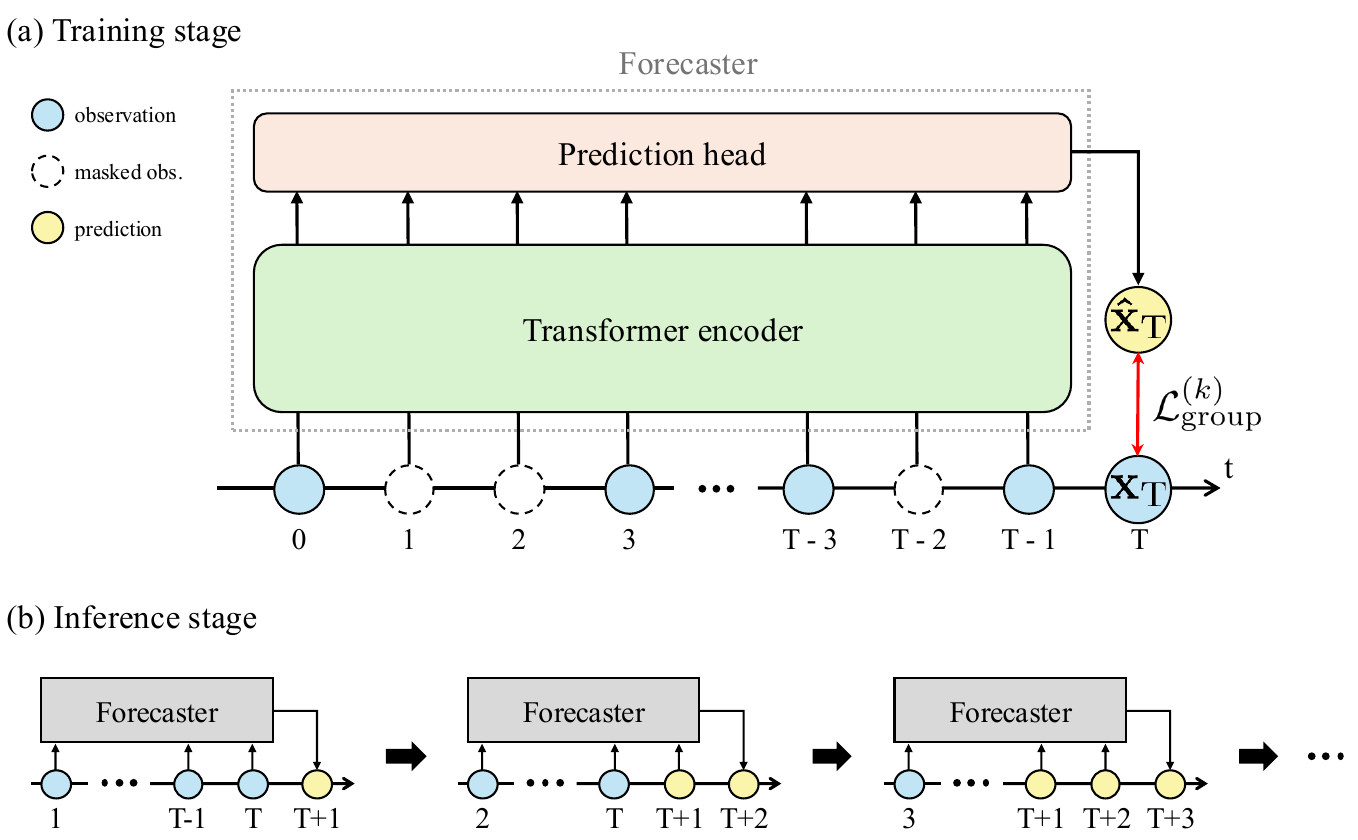}
\caption{\textbf{Overview of the forecaster.}
(a) \emph{Training stage:} The forecaster is trained for each Gaussian group~$\motiongroup{G}{k}$ by minimizing the loss~$\mathcal{L}_{\text{group}}^{(k)}$ between the predicted $\mathbf{\hat{x}}_t$ and the observed $\mathbf{x}_t$ at time~$T$. We randomly apply contiguous masking to the inputs and gradually reduce the masking ratio later in training.
(b) \emph{Inference stage:} Future motion is generated via autoregressive rollout for each $\motiongroup{G}{k}$, where the input sequence is iteratively updated with each newly predicted frame. Note that the input length is always $T\!-\!1$, the same as in training.}
\vspace{-4mm}
\label{sup_fig:forecaster}
\end{figure*}

\subsection{Motion-aware Gaussian Grouping}
\paragraph{Iterative Region Growing.}We implement our grouping strategy by modifying the official Gaga~\cite{lyu2024gaga} implementation. 
During initialization and keyframe-based seeding, we extract the closest 40\% front Gaussians with respect to the camera view. 
For the region-growing stage, each Gaussian is represented as a spatiotemporal feature vector 
$[\canonicalsub{\mu}{c,g},\,\bm{w}'_g]$, composed of its canonical-space mean and a PCA-reduced motion coefficient. 
We apply PCA to the motion coefficients of all dynamic Gaussians in 4DGS representation,  
scale the resulting PCA components by 0.3, and concatenate them with the Gaussian means.
Before performing region growing for each group, we determine the growing radius $\epsilon_r$ by computing 
the average neighbor distance among Gaussians belonging to the same group, ensuring that the expansion scale reflects 
the underlying local density.

\paragraph{Reassignment Strategy.}In addition, during the Gaussian relabeling step for unassigned Gaussians, each Gaussian updates its group label by 
voting among the 8 nearest group-assigned neighbors. To further enforce reliable and spatially consistent grouping, 
we then reassign only the closest 50\% of unassigned Gaussians, determined by their average distance to the neighboring 
group-assigned Gaussians.

\subsection{Group-wise Optimization}
Our group-wise optimization consists of two main stages: (i) rigid motion initialization, and (ii) group-wise refinement. 
In the first stage, we initialize rigid group motion parameters to enable our motion-aware objective. In the second stage, 
using the initialized motion information, we refine pre-optimized 4DGS representation to enforce stronger 
spatiotemporal consistency.

\paragraph{Rigid Motion Initialization.}
Before refining 4DGS representation, we first estimate the rigid body transformation parameters for rigid groups 
(i.e., groups with $\motiongroup{\tau}{k} = 1$) to implement our rigid motion anchoring loss. Specifically, for each 
rigid motion group, we compute a sequence of rigid transformations 
$\{\motiongroup{\bm{\Phi}}{k}_t = [\rigidmotion{R}{k}{c}{t} \mid \rigidmotion{t}{k}{c}{t}]\}_{t=1}^T$ 
by solving a Procrustes alignment problem. The alignment at timestep $t$ is given by
\begin{align}
\motiongroup{\bm{\Phi}}{k}_t 
&= \arg\min_{\rigidmotion{R}{k}{c}{t},\, \rigidmotion{t}{k}{c}{t}}
\sum_{g \in \motiongroup{G}{k}}
\left\|
\rigidmotion{R}{k}{c}{t}\,\canonicalsub{\mu}{c,g} 
+ \rigidmotion{t}{k}{c}{t}
- \deformed{\mu}{t,g}
\right\|_2^2,
\end{align}
where $\rigidmotion{R}{k}{c}{t} \in \mathrm{SO}(3)$ and $\rigidmotion{t}{k}{c}{t} \in \mathbb{R}^3$. 
Here, $\canonicalsub{\mu}{c,g}$ and $\deformed{\mu}{t,g}$ denote the canonical and deformed Gaussian centers.

\paragraph{Refinement of 4DGS Representation.}
Similar to the pre-optimization stage (\Cref{sup_sec:pre-optim}), our refinement is also implemented on top of 
SoM. We incorporate our group-wise motion optimization objective 
$\mathcal{L}_{\text{motion}}$ into the original training loss. For optimization stability, we apply scaling weights and 
reformulate the objective as:
\begin{equation}
    \mathcal{L}_{\text{motion}} 
    =
    \sum_{k=1}^{K} 
    \Big[
    \motiongroup{\tau}{k}\cdot \,\lambda_\text{rigid} \,\mathcal{L}_{\text{rigid}}^{(k)}
    +
    (1-\motiongroup{\tau}{k})\cdot \,\lambda_\text{nr}\,\mathcal{L}_{\text{nr}}^{(k)}
    \Big],
\end{equation}
where we set $\lambda_\text{rigid} = 0.1$ and $\lambda_\text{nr} = 0.02$ in all experiments.

We apply our constraint optimization objective together with the original reconstruction loss of SoM~\cite{wang2025shape}, defined as
\begin{equation}
    \mathcal{L}_{\text{recon}}
    =
    \|\hat{\mathbf{I}} - \mathbf{I}\|_{1}
    +
    \lambda_{\text{depth}}\|\hat{\mathbf{D}} - \mathbf{D}\|_{1}
    +
    \lambda_{\text{mask}}\|\hat{\mathbf{M}} - \mathbf{M}\|_{1},
\end{equation}
where $\hat{\mathbf{I}}, \hat{\mathbf{D}}, \hat{\mathbf{M}}$ denote the rendered RGB, depth, and mask predictions from current 4DGS representation.  
We follow the loss-weight configuration provided in the official SoM implementation.  
Our final optimization objective is then formulated as
\begin{equation}
    \mathcal{L}_{\text{total}}
    =
    \mathcal{L}_{\text{recon}}
    +
    \mathcal{L}_{\text{motion}}.
\end{equation}
This combined objective ensures that the 4DGS representation remains photometrically faithful while also enforcing coherent group-wise rigid and non-rigid motion behavior.

\begin{table*}[ht!]
\centering
\caption{\textbf{Forecasting results on the iPhone dataset.}
We forecast frames beyond the observed time window and evaluate the predicted frames rendered from held-out test viewpoints.
The leftmost column (\emph{Obs. ratio}) indicates the fraction of input training frames used by each model.
}
\vspace{-2mm}
\resizebox{\textwidth}{!}{
\begin{tabular}{llccccccccc}
\toprule
\multirow{2}{*}{\emph{Obs. ratio}} & \multirow{2}{*}{Scene} 
& \multicolumn{3}{c}{GSPred} 
& \multicolumn{3}{c}{GSPred-SoM} 
& \multicolumn{3}{c}{{\oursname} (Ours)} \\
\cmidrule(lr){3-5} \cmidrule(lr){6-8} \cmidrule(lr){9-11}
& 
& mPSNR$\uparrow$ & mSSIM$\uparrow$ & mLPIPS$\downarrow$
& mPSNR$\uparrow$ & mSSIM$\uparrow$ & mLPIPS$\downarrow$
& mPSNR$\uparrow$ & mSSIM$\uparrow$ & mLPIPS$\downarrow$ \\
\midrule
\multicolumn{1}{c}{\multirow{6}{*}{70\%}} 
& \multicolumn{1}{|l}{apple}
& \multicolumn{1}{|c}{13.21} & 0.5562 & 0.6942
& \multicolumn{1}{|c}{\textbf{16.58}} & \textbf{0.7339} & \textbf{0.5007}
& \multicolumn{1}{|c}{16.45} & 0.7220 & 0.5329 \\
& \multicolumn{1}{|l}{block}
& \multicolumn{1}{|c}{13.11} & 0.5333 & 0.5240
& \multicolumn{1}{|c}{13.47} & \textbf{0.6192} & \textbf{0.4900}
& \multicolumn{1}{|c}{\textbf{14.10}} & 0.6110 & 0.5030 \\
& \multicolumn{1}{|l}{paper-windmill}
& \multicolumn{1}{|c}{14.67} & 0.2168 & 0.4410
& \multicolumn{1}{|c}{18.73} & \textbf{0.5221} & 0.2191
& \multicolumn{1}{|c}{\textbf{18.98}} & 0.5197 & \textbf{0.2108} \\
& \multicolumn{1}{|l}{spin}
& \multicolumn{1}{|c}{15.67} & 0.4738 & 0.3975
& \multicolumn{1}{|c}{\textbf{16.00}} & \textbf{0.6797} & \textbf{0.3049}
& \multicolumn{1}{|c}{15.87} & 0.6618 & 0.3438 \\
& \multicolumn{1}{|l}{teddy}
& \multicolumn{1}{|c}{11.46} & \textbf{0.5886} & 0.5708
& \multicolumn{1}{|c}{10.84} & 0.5666 & 0.6555
& \multicolumn{1}{|c}{\textbf{11.87}} & 0.5562 & \textbf{0.5578} \\
\rowcolor{avgrow}
\multicolumn{1}{c}{\cellcolor{white}} 
& \multicolumn{1}{|l}{Average}
& \multicolumn{1}{|c}{13.62} & 0.4737 & 0.5255
& \multicolumn{1}{|c}{15.12} & \textbf{0.6243} & 0.4341
& \multicolumn{1}{|c}{\textbf{15.45}} & 0.6141 & \textbf{0.4296} \\
\midrule
\multicolumn{1}{c}{\multirow{6}{*}{90\%}} 
& \multicolumn{1}{|l}{apple}
& \multicolumn{1}{|c}{15.21} & 0.5550 & 0.4574
& \multicolumn{1}{|c}{16.88} & \textbf{0.8541} & \textbf{0.4453}
& \multicolumn{1}{|c}{\textbf{16.98}} & 0.8526 & 0.4530 \\
& \multicolumn{1}{|l}{block}
& \multicolumn{1}{|c}{\textbf{14.57}} & 0.5445 & \textbf{0.4909}
& \multicolumn{1}{|c}{12.49} & 0.5896 & 0.5363
& \multicolumn{1}{|c}{14.19} & \textbf{0.6018} & 0.4912 \\
& \multicolumn{1}{|l}{paper-windmill}
& \multicolumn{1}{|c}{14.78} & 0.2201 & 0.4280
& \multicolumn{1}{|c}{19.00} & \textbf{0.5559} & 0.2111
& \multicolumn{1}{|c}{\textbf{19.22}} & 0.5538 & \textbf{0.2036} \\
& \multicolumn{1}{|l}{spin}
& \multicolumn{1}{|c}{14.75} & 0.4631 & 0.4232
& \multicolumn{1}{|c}{16.14} & 0.6931 & 0.3146
& \multicolumn{1}{|c}{\textbf{16.99}} & \textbf{0.6975} & \textbf{0.2933} \\
& \multicolumn{1}{|l}{teddy}
& \multicolumn{1}{|c}{11.37} & \textbf{0.5816} & 0.5929
& \multicolumn{1}{|c}{11.99} & 0.5411 & 0.5894
& \multicolumn{1}{|c}{\textbf{12.39}} & 0.5433 & \textbf{0.5436} \\
\rowcolor{avgrow}
\multicolumn{1}{c}{\cellcolor{white}} 
& \multicolumn{1}{|l}{Average}
& \multicolumn{1}{|c}{14.13} & 0.4728 & 0.4785
& \multicolumn{1}{|c}{15.30} & 0.6467 & 0.4193
& \multicolumn{1}{|c}{\textbf{15.95}} & \textbf{0.6498} & \textbf{0.3970} \\
\bottomrule
\end{tabular}
}
\label{tab:iphone_70_90}
\vspace{-4mm}
\end{table*}

\subsection{Group-wise Motion Forecasting}
\paragraph{Architecture Details.}
We illustrate the architecture of our forecaster in Figure~\ref{sup_fig:forecaster}.
The model uses a single-layer Transformer encoder with a 32-dimensional hidden state, eight attention heads, and a 64-dimensional feedforward layer, followed by a flatten-head decoder. We apply a dropout rate of 0.2 and use instance normalization for each sequence. For both training and inference, we represent each Gaussian motion $\mathbf{x}_t$ at time~$t$ as
\begin{equation}
    \mathbf{x}_t = [\,\bm{\mu}_t,\, \mathbf{q}_t\,] \in \mathbb{R}^7,
\end{equation}
where $\bm{\mu}_t \in \mathbb{R}^3$ denotes the 3D translation vector and $\mathbf{q}_t \in \mathbb{R}^4$ denotes the unit quaternion that parameterizes the rotation.

\paragraph{Training Procedure and Inference Scheme.}
During training, we employ variable observation masking, where a contiguous span covering 40--0\% of the sequence is masked, and the masking ratio is gradually annealed over the course of training. To encourage physically plausible and smooth motion, we apply an acceleration regularization term with $\lambda_{\text{acc}}=1.0$ for all experiments. 

We further adopt a group-wise forecasting scheme in which one forecaster is trained per motion group, and each forecaster is responsible for training and predicting only the trajectories associated with its assigned group.

%% file: supple_sec/3_additional_results.tex
\section{Additional Evaluation Results}

\subsection{Future View Synthesis}

\paragraph{Evaluation on the iPhone Dataset.}
For the iPhone dataset~\cite{gao2022monocular}, we provide quantitative evaluation results for additional observation ratios (70\% and 90\%) in \Cref{tab:iphone_70_90}.  
We further present additional qualitative results for long-term forecasting (\Cref{sup_fig:iphone_spin_long_term_forecast}) and qualitative comparisons across diverse scenes (\Cref{sup_fig:qual_iphone}), extending \Cref{fig:main_iphone_qual}.

\paragraph{Evaluation on D-NeRF dataset.}
We further report evaluation results on D-NeRF dataset~\cite{pumarola2021d}. Using the same setup as GSPred~\cite{zhao2024gaussianprediction}, we use an 80\% observation ratio and a 20\% test split. In this setting, Gaussian motions are forecasted, and all trajectories are extrapolated using a weighted sum of the predicted motion. The quantitative results are summarized in Table~\ref{tab:dnerf_0.8}. Additionally, qualitative comparisons against GSPred are presented in Figure~\ref{fig:supple_dnerf_qual}.

\begin{table}[t!]
\centering
\caption{\textbf{Forecasting results on D-NeRF dataset.} We use 80\% of the frames as input and predict the remaining 20\%.}
\vspace{-2mm}
\resizebox{\columnwidth}{!}{
\begin{tabular}{lcccccc}
\toprule
\multirow{2}{*}{Scene} & \multicolumn{3}{c}{GSPred~\cite{zhao2024gaussianprediction}} & \multicolumn{3}{c}{Ours} \\
\cmidrule(lr){2-4} \cmidrule(lr){5-7}
 & PSNR$\uparrow$ & SSIM$\uparrow$ & LPIPS$\downarrow$
 & PSNR$\uparrow$ & SSIM$\uparrow$ & LPIPS$\downarrow$ \\
\midrule
\multicolumn{1}{l|}{Trex}            & \textbf{21.09} & \textbf{0.9406} & \multicolumn{1}{c|}{\textbf{0.0461}} & 20.60 & 0.9405 & 0.0499 \\
\multicolumn{1}{l|}{Mutant}          & 28.16 & \textbf{0.9560} & \multicolumn{1}{c|}{0.0256} & \textbf{28.72} & 0.9552 & \textbf{0.0218} \\
\multicolumn{1}{l|}{Jumpingjacks}    & \textbf{20.51} & 0.9184 & \multicolumn{1}{c|}{0.0760} & 20.12 & \textbf{0.9220} & \textbf{0.0734} \\
\multicolumn{1}{l|}{Standup}         & 25.96 & 0.9403 & \multicolumn{1}{c|}{0.0481} & \textbf{30.38} & \textbf{0.9544} & \textbf{0.0335} \\
\multicolumn{1}{l|}{Bouncingballs}   & 26.63 & 0.9714 & \multicolumn{1}{c|}{\textbf{0.0361}} & \textbf{26.82} & \textbf{0.9728} & 0.0367 \\
\multicolumn{1}{l|}{Hook}            & 23.42 & \textbf{0.9089} & \multicolumn{1}{c|}{0.0573} & \textbf{24.13} & 0.9081 & \textbf{0.0562} \\
\multicolumn{1}{l|}{Hellwarrior}     & 30.75 & 0.9281 & \multicolumn{1}{c|}{0.0729} & \textbf{31.29} & \textbf{0.9323} & \textbf{0.0680} \\
\multicolumn{1}{l|}{Lego}            & 11.99 & 0.7562 & \multicolumn{1}{c|}{0.2338} & \textbf{24.88} & \textbf{0.9005} & \textbf{0.0532} \\
\midrule
\multicolumn{1}{l|}{\textbf{Average}}& 23.56 & 0.9150 & \multicolumn{1}{c|}{0.0745} & \textbf{25.87} & \textbf{0.9357} & \textbf{0.0491} \\
\bottomrule
\end{tabular}
}
\label{tab:dnerf_0.8}
\vspace{-6mm}
\end{table}

\subsection{Novel View Synthesis}
To analyze the effect of our group-wise optimization on scene interpolation,  
we evaluate novel view synthesis (NVS) results and point-tracking performance on iPhone dataset.  
The experiment is conducted using the full training set (100\% observation ratio) so that the evaluation reflects pure interpolation performance.

As reported in \Cref{sup_tab:nvs_comparison}, our constrained optimization yields slight improvements in both point-tracking accuracy and photometric metrics.  
Qualitatively, \Cref{sup_fig:nvs_comparison} shows that our method enhances the geometric consistency of resulting 4DGS representation.
These results confirm that our Gaussian grouping and group-wise constrained optimization help obtain more physically plausible dynamic Gaussian representation.

\section{Component Analysis of {\oursname}}
\label{sup_sec:component-analysis}

\begin{table*}[ht!]
\centering
\caption{\textbf{Quantitative evaluation results on scene interpolation.}
We compare tracking and photometric performance against SoM~\cite{wang2025shape} on iPhone dataset~\cite{gao2022monocular}.}
\resizebox{0.75\textwidth}{!}{
\begin{tabular}{lccccccccc}
\toprule
\multirow{2}{*}{Methods} &
\multicolumn{3}{c}{3D tracking} &
\multicolumn{3}{c}{2D tracking} &
\multicolumn{3}{c}{Photometric} \\
\cmidrule(lr){2-4} \cmidrule(lr){5-7} \cmidrule(lr){8-10}
 & EPE$\downarrow$ & $\delta_{3D}^{.10}$$\uparrow$ & $\delta_{3D}^{.05}$$\uparrow$
 & AJ$\uparrow$ & $\delta_{\text{avg}}$$\uparrow$ & OA$\uparrow$
 & MPSNR$\uparrow$ & mSSIM$\uparrow$ & mLPIPS$\downarrow$ \\
\midrule
SoM~\cite{wang2025shape}
& 0.1016
& 70.62
& 46.59
& 37.61
& 49.17
& \textbf{86.99}
& 16.72
& \textbf{0.6462}
& 0.3914 \\
{\oursname} (Ours)
& \textbf{0.0989}
& \textbf{70.63}
& \textbf{47.64}
& \textbf{38.54}
& \textbf{49.93}
& 86.85
& \textbf{16.75}
& \textbf{0.6462}
& \textbf{0.3887} \\
\bottomrule
\end{tabular}
}
\vspace{-4mm}
\label{sup_tab:nvs_comparison}
\end{table*}

\begin{table}[t!]
\centering
\caption{\textbf{Ablation studies on group-wise design and region growing.} We analyze the contribution of each component in our method across 3D/2D motion tracking and novel view synthesis.}
\vspace{-2mm}

\begin{subtable}{\linewidth}
\centering
\caption{Effect of group-wise optimization, forecasting, and encoder capacity on future Gaussian motion forecasting. The gray row denotes MoGaF (Ours).}
\vspace{1mm}
\resizebox{\linewidth}{!}{
\begin{tabular}{c c c | cccccc}
    \toprule
    \multirow{2}{*}{Encoder} &
    \multicolumn{2}{c!{\color{black}\vrule}}{Group-wise} &
    \multicolumn{3}{c}{3D tracking} &
    \multicolumn{3}{c}{2D tracking} \\
    \cmidrule(lr){2-3} \cmidrule(lr){4-6} \cmidrule(lr){7-9}
     & Opt. & Forecast &
    EPE$\downarrow$ & $\delta_{3D}^{.10}$$\uparrow$ & $\delta_{3D}^{.05}$$\uparrow$ &
    AJ$\uparrow$ & $\delta_{\text{avg}}$$\uparrow$ & OA$\uparrow$ \\
    \midrule
    \multirow{4}{*}{\shortstack[c]{5-Layer\\(HC)}}
    & -- & -- & 0.253 & 37.5 & 12.7 & 12.8 & 7.9 & 75.9 \\
    & \checkmark & -- & 0.264 & 40.0 & 15.8 & 13.0 & 7.6 & 77.8 \\
    & -- & \checkmark & 0.246 & 42.6 & 16.7 & 16.0 & 8.5 & 74.8 \\
    & \checkmark & \checkmark & 0.245 & 43.5 & 19.5 & 16.1 & \textbf{9.4} & 79.2 \\
    \midrule
    \multirow{4}{*}{\shortstack[c]{1-Layer\\(LW)}}
    & -- & -- & 0.296 & 35.6 & 17.1 & 13.3 & 8.5 & 64.1 \\
    & \checkmark & -- & 0.269 & 36.7 & 15.8 & 13.4 & 7.6 & 70.5 \\
    & -- & \checkmark & 0.237 & 42.9 & 16.9 & 15.4 & 8.6 & 72.2 \\
    \rowcolor{avgrow}
    & \checkmark & \checkmark & \textbf{0.236} & \textbf{44.8} & \textbf{22.5} & \textbf{17.5} & 8.8 & \textbf{80.1} \\
    \bottomrule
\end{tabular}
}
\label{sup_tab:ablation_groupwise}
\end{subtable}

\vspace{3mm}

\begin{subtable}{\linewidth}
\centering
\caption{Forecasting results comparing a naive extension (Gaga-4D) and our motion-aware Gaussian grouping.}
\vspace{1mm}
\resizebox{1.0\linewidth}{!}{
\begin{tabular}{l cccccc}
\toprule
Method & mPSNR $\uparrow$ & mSSIM $\uparrow$ & mLPIPS $\downarrow$ & $\delta^{.10}_{3D}$ $\uparrow$ & $\delta^{.05}_{3D}$ $\uparrow$ & OA $\uparrow$\\
\midrule
Gaga-4D
& 15.31 & 0.6015 & 0.4475 & 43.4 & 20.9 & 75.6 \\
Ours
& \textbf{15.51} & \textbf{0.6143} & \textbf{0.4245} & \textbf{44.8} & \textbf{22.5} & \textbf{80.1} \\
\bottomrule
\end{tabular}
}
\label{sup_tab:grouping_ablation}
\end{subtable}

\vspace{-4mm}
\end{table}

For a deeper analysis of our pipeline, we conduct a component-wise analysis of {\oursname} and evaluate the effectiveness of our motion-aware Gaussian grouping strategy.  
The component analysis focuses on three key design axes:  
(1) group-wise optimization,  
(2) the group-wise forecasting scheme, and  
(3) the capacity of the motion forecaster.  
For the grouping strategy, we compare our method with a simple extension of static Gaussian grouping~\cite{lyu2024gaga}. All experiments are conducted with a 60\% observation ratio, forecasting the remaining 40\% of each sequence on iPhone dataset~\cite{gao2022monocular}.

\subsection{Analysis for Group-wise Design}
\label{sup_sec:group-optim-analysis}

To isolate the impact of each component, we disable the group-wise optimization and group-wise forecasting modules individually.  
We further evaluate a higher-capacity motion forecaster—a 5-layer transformer encoder (HC)—against our default 1-layer transformer (LW) encoder. 
To accurately evaluate forecasting performance, we compare various configurations using point tracking accuracy.  

\paragraph{Effect of Group-wise Optimization.}
\Cref{sup_tab:ablation_groupwise} summarizes the results of our component analysis.
For both HC and LW forecasters, enabling group-wise optimization consistently improves 3D tracking accuracy.  
This suggests that object-level motion constraints help preserve coherent and stable spatiotemporal trajectories within 4DGS representation.

\paragraph{Contribution of Group-wise Forecasting.}
Group-wise forecasting further improves performance across all metrics.  
By learning motion patterns within each object independently, the forecaster produces more stable long-term predictions and higher 2D tracking accuracy.  
When combined with group-wise optimization, it yields the best results for both HC and LW settings, highlighting the complementary roles of the two modules.

\paragraph{Effect of Forecaster Capacity.}
Interestingly, the HC forecaster does not provide noticeable improvement over the LW version.
We hypothesize that larger models may be more sensitive to noisy or frame-specific motion variations in 4DGS, tend to introduce instability in long-term predictions. 
In contrast, the lightweight forecaster appears to leverage the structured, object-level motion units more effectively,  
suggesting that strong forecasting performance is achievable without large model capacity.

\subsection{Effect of Motion-aware Gaussian Grouping}
\label{sup_sec:grouping-analysis}

We evaluate the effectiveness of our motion-aware Gaussian grouping strategy for scene forecasting.  
Specifically, we compare our approach with a baseline that applies iterative region growing based on a naive extension of static Gaussian grouping (Gaga-4D), which is described in \Cref{sup_sec:gaga4D}.

As shown in \Cref{sup_tab:grouping_ablation}, our method consistently outperforms the baseline across all evaluation metrics, including photometric metrics and point tracking accuracy.
These results indicate that our motion-aware grouping enables more accurate separation of Gaussians with coherent motion, leading to improved extrapolation under out-of-distribution (OOD) conditions.  
Overall, these results demonstrate that our region growing strategy effectively enhances motion awareness in Gaussian grouping under complex dynamic scenes.

%% file: supple_sec/4_limitations.tex
\section{Limitations and Future Work}

Our method exhibits several limitations arising from its dependence on pretrained 4DGS representation.
Since the grouping, optimization, and forecasting stages operate on geometry produced by the baseline 4DGS, the overall quality is bounded by the fidelity of this reconstruction.
Thus, the method cannot recover unseen geometry, and severe failures in the underlying motion optimization—especially in highly complex scenes—directly propagate to later stages.
Incorporating point-cloud completion techniques or generative priors may alleviate this dependency by providing additional geometric constraints.

Another limitation appears when object groups overlap during forecasting.
Because Gaussians are not physical primitives, our model cannot reason about collisions or inter-object interactions, offering limited ability to detect or prevent physically implausible overlap between motion groups.
Extending the framework with physical constraints or physically grounded Gaussian representations would be a promising direction for future work.
\clearpage

%% file: supple_sec/5_figures.tex
\begin{figure*}[ht!]
\centering
\includegraphics[width=0.65\textwidth]{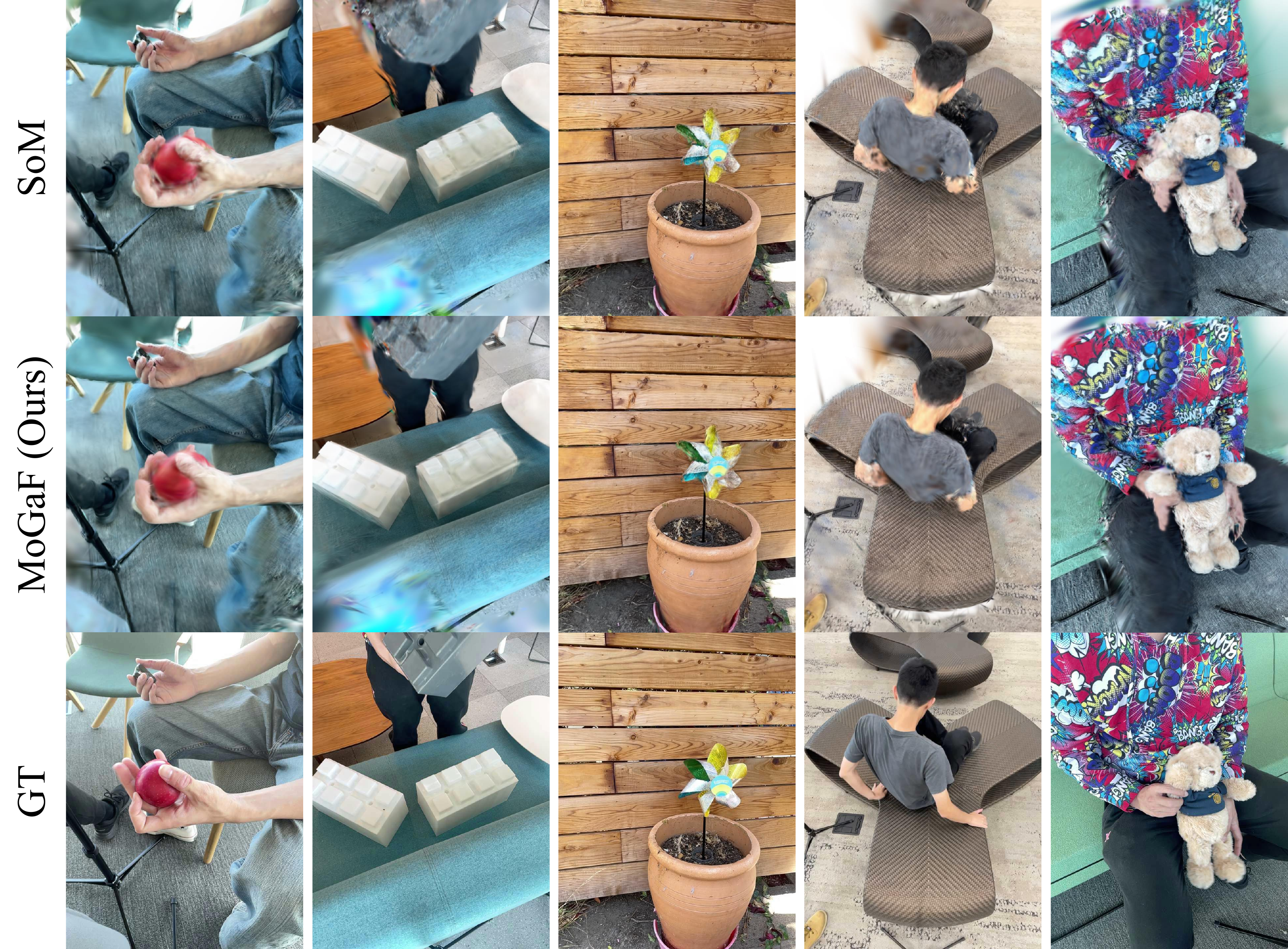}
\vspace{-2mm}
\caption{\textbf{Qualitative evaluation results on scene interpolation.}
We report NVS performance of SoM~\cite{wang2025shape} and {\oursname} on iPhone dataset~\cite{gao2022monocular}, evaluated on test camera viewpoints over the full training sequence.}
\label{sup_fig:nvs_comparison}
\vspace{-2.5mm}
\end{figure*}

\begin{figure*}[ht!]
\centering
\includegraphics[width=\textwidth]{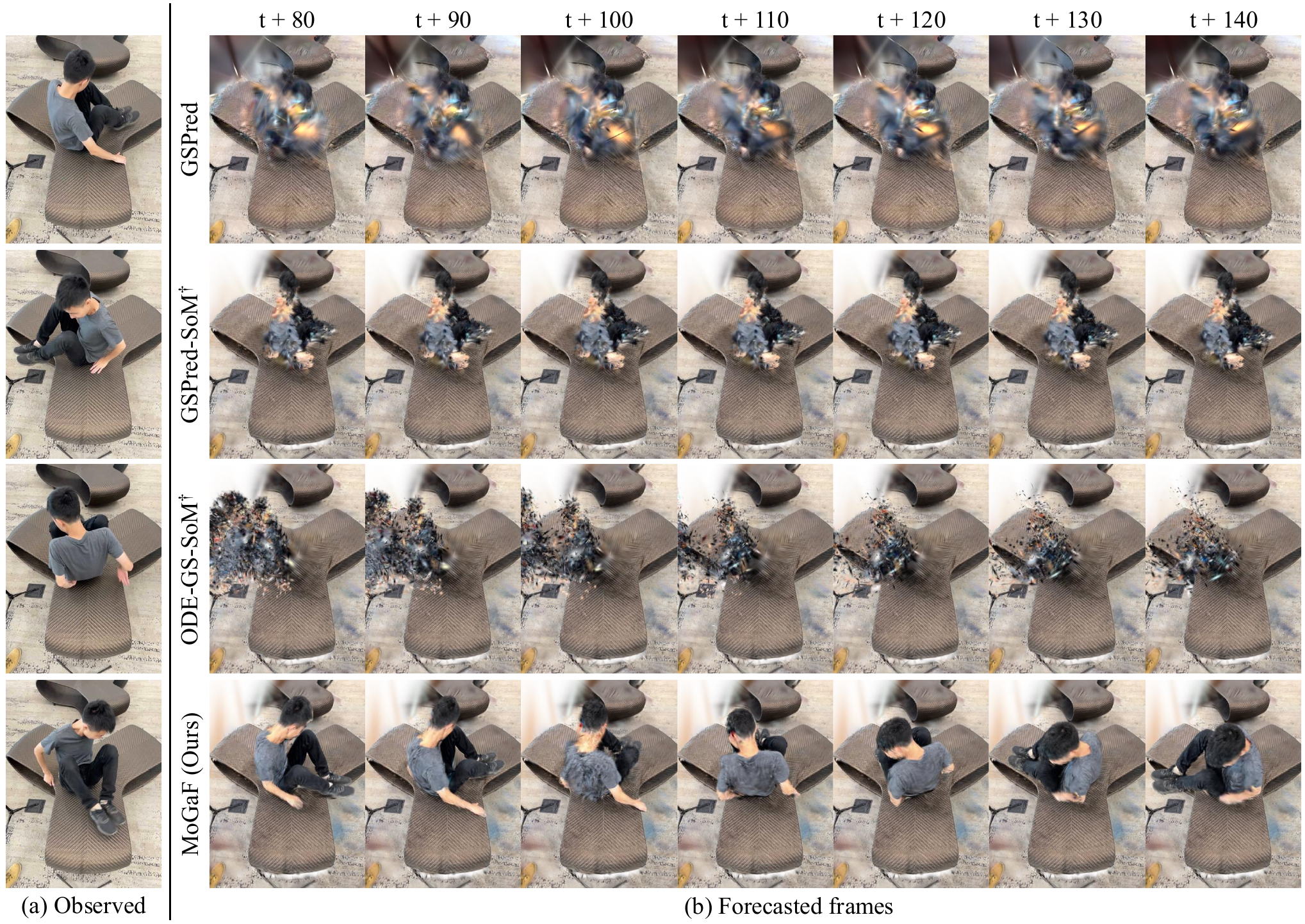}
\vspace{-7mm}
\caption{\textbf{Long-term forecasting results on iPhone dataset.}
(a) shows GT views from test cameras at the observed timesteps, and (b) presents the forecasted renderings for timesteps beyond the observations. Note that $t$ denotes the timestep of the last observed frame.}
\label{sup_fig:iphone_spin_long_term_forecast}
\end{figure*}

\begin{figure*}[ht!]
\centering
\includegraphics[width=\textwidth]{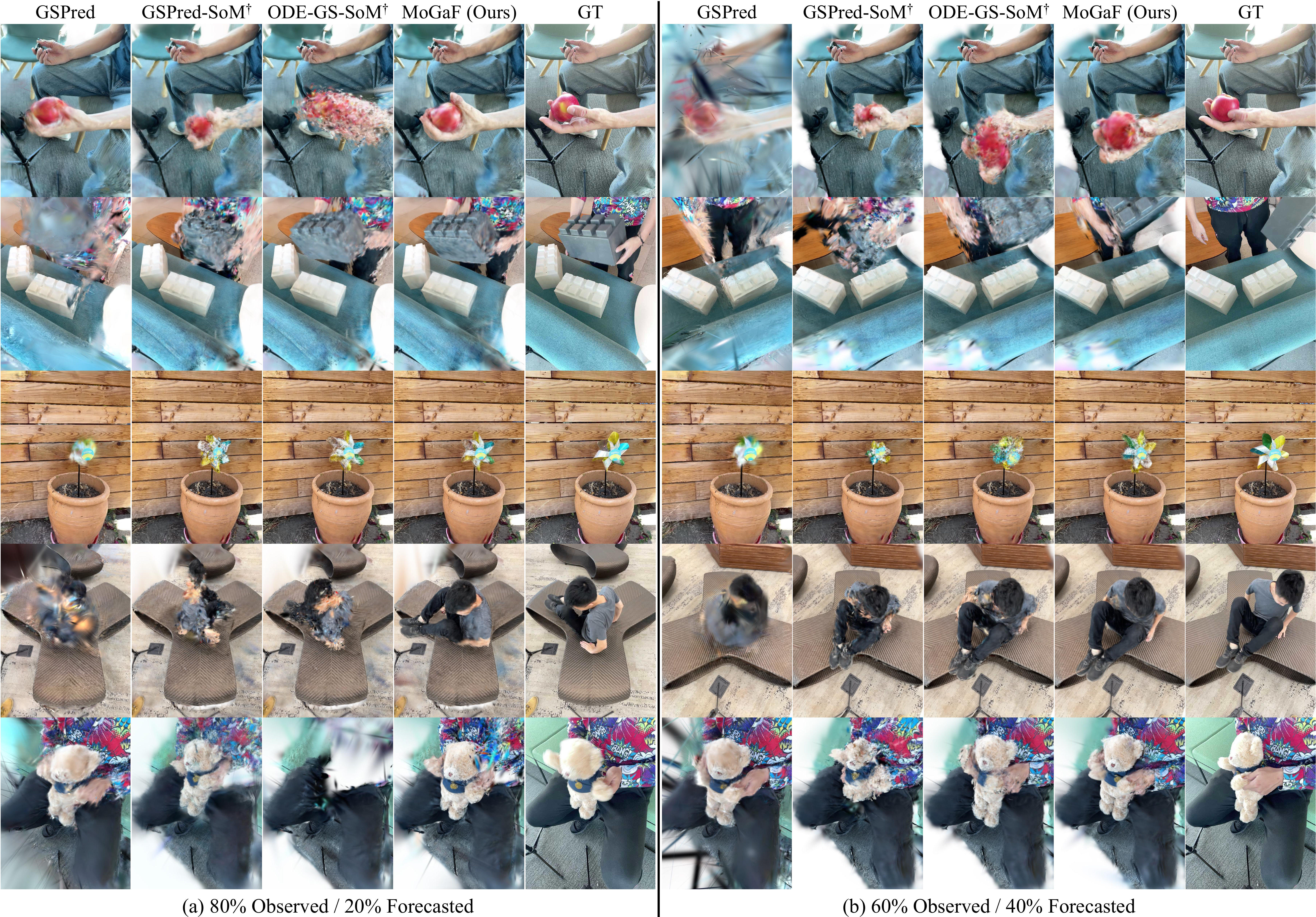}
\caption{\textbf{Qualitative results on iPhone dataset.}
We present forecasted frames from test camera views. 
(a) and (b) correspond to settings where the first 80\% and 60\% of frames are used for training, and the remaining 20\% and 40\% are forecasted, respectively.}
\label{sup_fig:qual_iphone}
\end{figure*}

\begin{figure*}[ht!]
\centering
\includegraphics[width=\textwidth]{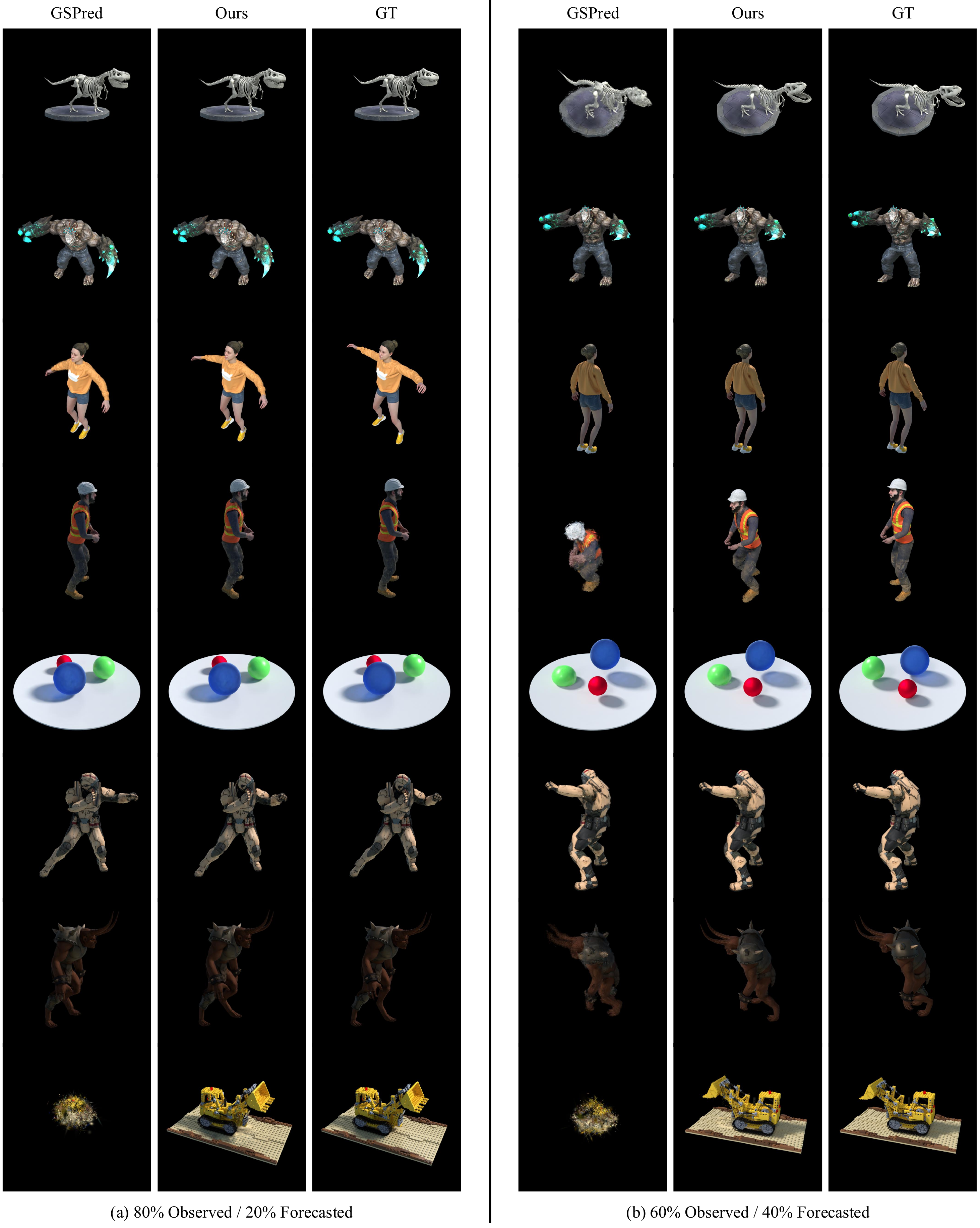}
\caption{\textbf{Qualitative results on D-NeRF dataset.}
We present forecasted frames from test camera views. 
(a) and (b) correspond to settings where the first 80\% and 60\% of frames are used for training, and the remaining 20\% and 40\% are forecasted, respectively.}
\label{fig:supple_dnerf_qual}
\end{figure*}

%% file: main.bib
@String(ECCV= {Eur. Conf. Comput. Vis.})

@String(ICPR = {Int. Conf. Pattern Recog.})

@String(ICIP = {IEEE Int. Conf. Image Process.})

@String(ECCV  = {ECCV})

@String(ICPR  = {ICPR})

@String(ICIP  = {ICIP})

@inproceedings{zhao2024gaussianprediction,
  title={Gaussianprediction: Dynamic 3d gaussian prediction for motion extrapolation and free view synthesis},
  author={Zhao, Boming and Li, Yuan and Sun, Ziyu and Zeng, Lin and Shen, Yujun and Ma, Rui and Zhang, Yinda and Bao, Hujun and Cui, Zhaopeng},
  booktitle={ACM SIGGRAPH 2024 Conference Papers},
  pages={1--12},
  year={2024}
}

@article{wang2025ode,
  title={Ode-gs: Latent odes for dynamic scene extrapolation with 3d gaussian splatting},
  author={Wang, Daniel and Rim, Patrick and Tian, Tian and Lao, Dong and Wong, Alex and Sundaramoorthi, Ganesh},
  journal={arXiv preprint arXiv:2506.05480},
  year={2025}
}

@inproceedings{wang2025shape,
  title={Shape of motion: 4d reconstruction from a single video},
  author={Wang, Qianqian and Ye, Vickie and Gao, Hang and Zeng, Weijia and Austin, Jake and Li, Zhengqi and Kanazawa, Angjoo},
  booktitle={Proceedings of the IEEE/CVF International Conference on Computer Vision},
  pages={9660--9672},
  year={2025}
}

@ARTICLE{ewasplatting,
  author={Zwicker, M. and Pfister, H. and van Baar, J. and Gross, M.},
  journal={IEEE Transactions on Visualization and Computer Graphics}, 
  title={EWA splatting}, 
  year={2002},
  volume={8},
  number={3},
  pages={223-238},
  doi={10.1109/TVCG.2002.1021576}}

@article{kerbl20233d,
  title={3d gaussian splatting for real-time radiance field rendering.},
  author={Kerbl, Bernhard and Kopanas, Georgios and Leimk{\"u}hler, Thomas and Drettakis, George},
  journal={ACM Trans. Graph.},
  volume={42},
  number={4},
  pages={139--1},
  year={2023}
}

@article{lyu2024gaga,
  title={Gaga: Group any gaussians via 3d-aware memory bank},
  author={Lyu, Weijie and Li, Xueting and Kundu, Abhijit and Tsai, Yi-Hsuan and Yang, Ming-Hsuan},
  journal={arXiv preprint arXiv:2404.07977},
  year={2024}
}

@inproceedings{ye2024gaussian,
  title={Gaussian grouping: Segment and edit anything in 3d scenes},
  author={Ye, Mingqiao and Danelljan, Martin and Yu, Fisher and Ke, Lei},
  booktitle={European conference on computer vision},
  pages={162--179},
  year={2024},
  organization={Springer}
}

@misc{ren2024grounding,
      title={Grounding DINO 1.5: Advance the "Edge" of Open-Set Object Detection}, 
      author={Tianhe Ren and Qing Jiang and Shilong Liu and Zhaoyang Zeng and Wenlong Liu and Han Gao and Hongjie Huang and Zhengyu Ma and Xiaoke Jiang and Yihao Chen and Yuda Xiong and Hao Zhang and Feng Li and Peijun Tang and Kent Yu and Lei Zhang},
      year={2024},
      eprint={2405.10300},
      archivePrefix={arXiv},
      primaryClass={cs.CV}
}

@inproceedings{pumarola2021d,
  title={D-nerf: Neural radiance fields for dynamic scenes},
  author={Pumarola, Albert and Corona, Enric and Pons-Moll, Gerard and Moreno-Noguer, Francesc},
  booktitle={Proceedings of the IEEE/CVF conference on computer vision and pattern recognition},
  pages={10318--10327},
  year={2021}
}

@inproceedings{park2021nerfies,
  title={Nerfies: Deformable neural radiance fields},
  author={Park, Keunhong and Sinha, Utkarsh and Barron, Jonathan T and Bouaziz, Sofien and Goldman, Dan B and Seitz, Steven M and Martin-Brualla, Ricardo},
  booktitle={Proceedings of the IEEE/CVF international conference on computer vision},
  pages={5865--5874},
  year={2021}
}

@article{park2021hypernerf,
  title={Hypernerf: A higher-dimensional representation for topologically varying neural radiance fields},
  author={Park, Keunhong and Sinha, Utkarsh and Hedman, Peter and Barron, Jonathan T and Bouaziz, Sofien and Goldman, Dan B and Martin-Brualla, Ricardo and Seitz, Steven M},
  journal={arXiv preprint arXiv:2106.13228},
  year={2021}
}

@inproceedings{liu2018dyan,
  title={Dyan: A dynamical atoms-based network for video prediction},
  author={Liu, Wenqian and Sharma, Abhishek and Camps, Octavia and Sznaier, Mario},
  booktitle={Proceedings of the European Conference on Computer Vision (ECCV)},
  pages={170--185},
  year={2018}
}

@article{yan2021videogpt,
  title={Videogpt: Video generation using vq-vae and transformers},
  author={Yan, Wilson and Zhang, Yunzhi and Abbeel, Pieter and Srinivas, Aravind},
  journal={arXiv preprint arXiv:2104.10157},
  year={2021}
}

@inproceedings{ye2022vptr,
  title={Vptr: Efficient transformers for video prediction},
  author={Ye, Xi and Bilodeau, Guillaume-Alexandre},
  booktitle={2022 26th International conference on pattern recognition (ICPR)},
  pages={3492--3499},
  year={2022},
  organization={IEEE}
}

@inproceedings{wang2018eidetic,
    title={Eidetic 3D {LSTM}: A Model for Video Prediction and Beyond},
    author={Yunbo Wang and Lu Jiang and Ming-Hsuan Yang and Li-Jia Li and Mingsheng Long and Li Fei-Fei},
    booktitle={International Conference on Learning Representations},
    year={2019},
    url={https://openreview.net/forum?id=B1lKS2AqtX},
}

@article{gupta2022maskvit,
  title={Maskvit: Masked visual pre-training for video prediction},
  author={Gupta, Agrim and Tian, Stephen and Zhang, Yunzhi and Wu, Jiajun and Mart{\'\i}n-Mart{\'\i}n, Roberto and Fei-Fei, Li},
  journal={arXiv preprint arXiv:2206.11894},
  year={2022}
}

@inproceedings{villar2023object,
  title={Object-centric video prediction via decoupling of object dynamics and interactions},
  author={Villar-Corrales, Angel and Wahdan, Ismail and Behnke, Sven},
  booktitle={2023 IEEE International Conference on Image Processing (ICIP)},
  pages={570--574},
  year={2023},
  organization={IEEE}
}

@article{vaswani2017attention,
  title={Attention is all you need},
  author={Vaswani, Ashish and Shazeer, Noam and Parmar, Niki and Uszkoreit, Jakob and Jones, Llion and Gomez, Aidan N and Kaiser, {\L}ukasz and Polosukhin, Illia},
  journal={Advances in neural information processing systems},
  volume={30},
  year={2017}
}

@inproceedings{devlin2019bert,
  title={Bert: Pre-training of deep bidirectional transformers for language understanding},
  author={Devlin, Jacob and Chang, Ming-Wei and Lee, Kenton and Toutanova, Kristina},
  booktitle={Proceedings of the 2019 conference of the North American chapter of the association for computational linguistics: human language technologies, volume 1 (long and short papers)},
  pages={4171--4186},
  year={2019}
}

@inproceedings{lei2025mosca,
  title={Mosca: Dynamic gaussian fusion from casual videos via 4d motion scaffolds},
  author={Lei, Jiahui and Weng, Yijia and Harley, Adam W and Guibas, Leonidas and Daniilidis, Kostas},
  booktitle={Proceedings of the Computer Vision and Pattern Recognition Conference},
  pages={6165--6177},
  year={2025}
}

@article{gao2022monocular,
  title={Monocular dynamic view synthesis: A reality check},
  author={Gao, Hang and Li, Ruilong and Tulsiani, Shubham and Russell, Bryan and Kanazawa, Angjoo},
  journal={Advances in Neural Information Processing Systems},
  volume={35},
  pages={33768--33780},
  year={2022}
}

@article{ravi2024sam,
  title={Sam 2: Segment anything in images and videos},
  author={Ravi, Nikhila and Gabeur, Valentin and Hu, Yuan-Ting and Hu, Ronghang and Ryali, Chaitanya and Ma, Tengyu and Khedr, Haitham and R{\"a}dle, Roman and Rolland, Chloe and Gustafson, Laura and others},
  journal={arXiv preprint arXiv:2408.00714},
  year={2024}
}

@article{mildenhall2021nerf,
  title={Nerf: Representing scenes as neural radiance fields for view synthesis},
  author={Mildenhall, Ben and Srinivasan, Pratul P and Tancik, Matthew and Barron, Jonathan T and Ramamoorthi, Ravi and Ng, Ren},
  journal={Communications of the ACM},
  volume={65},
  number={1},
  pages={99--106},
  year={2021},
  publisher={ACM New York, NY, USA}
}

@inproceedings{li2022neural,
  title={Neural 3d video synthesis from multi-view video},
  author={Li, Tianye and Slavcheva, Mira and Zollhoefer, Michael and Green, Simon and Lassner, Christoph and Kim, Changil and Schmidt, Tanner and Lovegrove, Steven and Goesele, Michael and Newcombe, Richard and others},
  booktitle={Proceedings of the IEEE/CVF conference on computer vision and pattern recognition},
  pages={5521--5531},
  year={2022}
}

@inproceedings{fridovich2023k,
  title={K-planes: Explicit radiance fields in space, time, and appearance},
  author={Fridovich-Keil, Sara and Meanti, Giacomo and Warburg, Frederik Rahb{\ae}k and Recht, Benjamin and Kanazawa, Angjoo},
  booktitle={Proceedings of the IEEE/CVF Conference on Computer Vision and Pattern Recognition},
  pages={12479--12488},
  year={2023}
}

@inproceedings{cao2023hexplane,
  title={Hexplane: A fast representation for dynamic scenes},
  author={Cao, Ang and Johnson, Justin},
  booktitle={Proceedings of the IEEE/CVF Conference on Computer Vision and Pattern Recognition},
  pages={130--141},
  year={2023}
}

@inproceedings{luiten2024dynamic,
  title={Dynamic 3d gaussians: Tracking by persistent dynamic view synthesis},
  author={Luiten, Jonathon and Kopanas, Georgios and Leibe, Bastian and Ramanan, Deva},
  booktitle={2024 International Conference on 3D Vision (3DV)},
  pages={800--809},
  year={2024},
  organization={IEEE}
}

@inproceedings{wu20244d,
  title={4d gaussian splatting for real-time dynamic scene rendering},
  author={Wu, Guanjun and Yi, Taoran and Fang, Jiemin and Xie, Lingxi and Zhang, Xiaopeng and Wei, Wei and Liu, Wenyu and Tian, Qi and Wang, Xinggang},
  booktitle={Proceedings of the IEEE/CVF conference on computer vision and pattern recognition},
  pages={20310--20320},
  year={2024}
}

@inproceedings{kratimenos2024dynmf,
  title={Dynmf: Neural motion factorization for real-time dynamic view synthesis with 3d gaussian splatting},
  author={Kratimenos, Agelos and Lei, Jiahui and Daniilidis, Kostas},
  booktitle={European Conference on Computer Vision},
  pages={252--269},
  year={2024},
  organization={Springer}
}

@article{yang2023real,
  title={Real-time photorealistic dynamic scene representation and rendering with 4d gaussian splatting},
  author={Yang, Zeyu and Yang, Hongye and Pan, Zijie and Zhang, Li},
  journal={arXiv preprint arXiv:2310.10642},
  year={2023}
}

@inproceedings{yang2024deformable,
  title={Deformable 3d gaussians for high-fidelity monocular dynamic scene reconstruction},
  author={Yang, Ziyi and Gao, Xinyu and Zhou, Wen and Jiao, Shaohui and Zhang, Yuqing and Jin, Xiaogang},
  booktitle={Proceedings of the IEEE/CVF conference on computer vision and pattern recognition},
  pages={20331--20341},
  year={2024}
}

@inproceedings{yang2024depth,
  title={Depth anything: Unleashing the power of large-scale unlabeled data},
  author={Yang, Lihe and Kang, Bingyi and Huang, Zilong and Xu, Xiaogang and Feng, Jiashi and Zhao, Hengshuang},
  booktitle={Proceedings of the IEEE/CVF conference on computer vision and pattern recognition},
  pages={10371--10381},
  year={2024}
}

@article{yang2023track,
  title={Track anything: Segment anything meets videos},
  author={Yang, Jinyu and Gao, Mingqi and Li, Zhe and Gao, Shang and Wang, Fangjing and Zheng, Feng},
  journal={arXiv preprint arXiv:2304.11968},
  year={2023}
}

@inproceedings{karaev2024cotracker,
  title={Cotracker: It is better to track together},
  author={Karaev, Nikita and Rocco, Ignacio and Graham, Benjamin and Neverova, Natalia and Vedaldi, Andrea and Rupprecht, Christian},
  booktitle={European conference on computer vision},
  pages={18--35},
  year={2024},
  organization={Springer}
}

@inproceedings{doersch2023tapir,
  title={Tapir: Tracking any point with per-frame initialization and temporal refinement},
  author={Doersch, Carl and Yang, Yi and Vecerik, Mel and Gokay, Dilara and Gupta, Ankush and Aytar, Yusuf and Carreira, Joao and Zisserman, Andrew},
  booktitle={Proceedings of the IEEE/CVF International Conference on Computer Vision},
  pages={10061--10072},
  year={2023}
}

@article{cen2025segment,
  title={Segment anything in 3d with radiance fields},
  author={Cen, Jiazhong and Fang, Jiemin and Zhou, Zanwei and Yang, Chen and Xie, Lingxi and Zhang, Xiaopeng and Shen, Wei and Tian, Qi},
  journal={International Journal of Computer Vision},
  pages={1--23},
  year={2025},
  publisher={Springer}
}

@misc{luo2025instant4d4dgaussiansplatting,
      title={Instant4D: 4D Gaussian Splatting in Minutes}, 
      author={Zhanpeng Luo and Haoxi Ran and Li Lu},
      year={2025},
      eprint={2510.01119},
      archivePrefix={arXiv},
      primaryClass={cs.CV},
      url={https://arxiv.org/abs/2510.01119}, 
}

@inproceedings{kirillov2023segment,
  title={Segment anything},
  author={Kirillov, Alexander and Mintun, Eric and Ravi, Nikhila and Mao, Hanzi and Rolland, Chloe and Gustafson, Laura and Xiao, Tete and Whitehead, Spencer and Berg, Alexander C and Lo, Wan-Yen and others},
  booktitle={Proceedings of the IEEE/CVF international conference on computer vision},
  pages={4015--4026},
  year={2023}
}

@article{chen2023interactive,
  title={Interactive segment anything nerf with feature imitation},
  author={Chen, Xiaokang and Tang, Jiaxiang and Wan, Diwen and Wang, Jingbo and Zeng, Gang},
  journal={arXiv preprint arXiv:2305.16233},
  year={2023}
}

@inproceedings{kim2024garfield,
  title={Garfield: Group anything with radiance fields},
  author={Kim, Chung Min and Wu, Mingxuan and Kerr, Justin and Goldberg, Ken and Tancik, Matthew and Kanazawa, Angjoo},
  booktitle={Proceedings of the IEEE/CVF Conference on Computer Vision and Pattern Recognition},
  pages={21530--21539},
  year={2024}
}

@inproceedings{zhou2024feature,
  title={Feature 3dgs: Supercharging 3d gaussian splatting to enable distilled feature fields},
  author={Zhou, Shijie and Chang, Haoran and Jiang, Sicheng and Fan, Zhiwen and Zhu, Zehao and Xu, Dejia and Chari, Pradyumna and You, Suya and Wang, Zhangyang and Kadambi, Achuta},
  booktitle={Proceedings of the IEEE/CVF Conference on Computer Vision and Pattern Recognition},
  pages={21676--21685},
  year={2024}
}

@inproceedings{cheng2023tracking,
  title={Tracking anything with decoupled video segmentation},
  author={Cheng, Ho Kei and Oh, Seoung Wug and Price, Brian and Schwing, Alexander and Lee, Joon-Young},
  booktitle={Proceedings of the IEEE/CVF International Conference on Computer Vision},
  pages={1316--1326},
  year={2023}
}

@inproceedings{dou2024learning,
  title={Learning segmented 3D Gaussians via efficient feature unprojection for zero-shot neural scene segmentation},
  author={Dou, Bin and Zhang, Tianyu and Wang, Zhaohui and Ma, Yongjia and Yuan, Zejian and Zheng, Nanning},
  booktitle={International Conference on Neural Information Processing},
  pages={398--412},
  year={2024},
  organization={Springer}
}

@article{ming2024survey,
  title={A survey on future frame synthesis: Bridging deterministic and generative approaches},
  author={Ming, Ruibo and Huang, Zhewei and Ju, Zhuoxuan and Hu, Jianming and Peng, Lihui and Zhou, Shuchang},
  journal={arXiv preprint arXiv:2401.14718},
  year={2024}
}

@inproceedings{wu2020future,
  title={Future video synthesis with object motion prediction},
  author={Wu, Yue and Gao, Rongrong and Park, Jaesik and Chen, Qifeng},
  booktitle={Proceedings of the IEEE/CVF Conference on Computer Vision and Pattern Recognition},
  pages={5539--5548},
  year={2020}
}

@inproceedings{wu2022optimizing,
  title={Optimizing video prediction via video frame interpolation},
  author={Wu, Yue and Wen, Qiang and Chen, Qifeng},
  booktitle={Proceedings of the IEEE/CVF Conference on Computer Vision and Pattern Recognition},
  pages={17814--17823},
  year={2022}
}

@inproceedings{girdhar2021anticipative,
  title={Anticipative video transformer},
  author={Girdhar, Rohit and Grauman, Kristen},
  booktitle={Proceedings of the IEEE/CVF international conference on computer vision},
  pages={13505--13515},
  year={2021}
}

@inproceedings{kwon2019predicting,
  title={Predicting future frames using retrospective cycle gan},
  author={Kwon, Yong-Hoon and Park, Min-Gyu},
  booktitle={Proceedings of the IEEE/CVF Conference on Computer Vision and Pattern Recognition},
  pages={1811--1820},
  year={2019}
}

@inproceedings{lu2017flexible,
  title={Flexible spatio-temporal networks for video prediction},
  author={Lu, Chaochao and Hirsch, Michael and Scholkopf, Bernhard},
  booktitle={Proceedings of the IEEE Conference on Computer Vision and Pattern Recognition},
  pages={6523--6531},
  year={2017}
}

@article{wu2024ivideogpt,
  title={ivideogpt: Interactive videogpts are scalable world models},
  author={Wu, Jialong and Yin, Shaofeng and Feng, Ningya and He, Xu and Li, Dong and Hao, Jianye and Long, Mingsheng},
  journal={Advances in Neural Information Processing Systems},
  volume={37},
  pages={68082--68119},
  year={2024}
}

@article{assran2025v,
  title={V-jepa 2: Self-supervised video models enable understanding, prediction and planning},
  author={Assran, Mido and Bardes, Adrien and Fan, David and Garrido, Quentin and Howes, Russell and Muckley, Matthew and Rizvi, Ammar and Roberts, Claire and Sinha, Koustuv and Zholus, Artem and others},
  journal={arXiv preprint arXiv:2506.09985},
  year={2025}
}

@article{hafner2019dream,
  title={Dream to control: Learning behaviors by latent imagination},
  author={Hafner, Danijar and Lillicrap, Timothy and Ba, Jimmy and Norouzi, Mohammad},
  journal={arXiv preprint arXiv:1912.01603},
  year={2019}
}
